\documentclass[letterpaper]{article} 
\usepackage{aaai23}  
\usepackage{times}  
\usepackage{helvet}  
\usepackage{courier}  
\usepackage[hyphens]{url}  
\usepackage{graphicx} 
\usepackage{natbib}  
\usepackage{caption} 
\usepackage{algorithm}
\usepackage{algorithmic}

\usepackage{graphicx} 
\usepackage{subcaption}

\usepackage{courier}

\usepackage{makecell}
\usepackage{color}
\usepackage{amsfonts}
\usepackage{amsmath}
\usepackage{amssymb}

\def\ie{\textit{i.e.}}
\def\eg{\textit{e.g.}}

\def\mD{{\mathcal D}}

\def\mS{{\mathcal S}}
\def\mT{{\mathcal T}}

\DeclareMathAlphabet\mathbfcal{OMS}{cmsy}{b}{n}

\def\0{{\bf 0}}
\def\1{{\bf 1}}

\def\eg{\emph{e.g.,}} 
\def\ie{\emph{i.e.,}} 
 
\def\etc{\emph{etc.}} \def\vs{\emph{vs.}}

\def\na{N/A}

\usepackage{booktabs,multirow,array,color}
\usepackage{xspace}
\usepackage{subcaption}
\usepackage{amsmath}
\usepackage{amsthm}
\usepackage{tikz}
\usepackage{amsfonts}
\usepackage{makecell}
\usepackage{enumitem}
\usepackage{colortbl}
\usepackage{bbding}
\usepackage{newfloat}
\usepackage{listings}
\DeclareCaptionStyle{ruled}{labelfont=normalfont,labelsep=colon,strut=off} 
\floatstyle{ruled}
\newfloat{listing}{tb}{lst}{}
\floatname{listing}{Listing}
\title{Instance Segmentation for Chinese Character Stroke Extraction, \\Datasets and Benchmarks}
\author {
    Lizhao Liu\textsuperscript{\rm 1}\footnote{This work was
partially done while the author was an intern at Tencent Cloud Xiaowei.},
    Kunyang Lin\textsuperscript{\rm 1},
    Shangxin Huang\textsuperscript{\rm 1},
    Zhongli Li\textsuperscript{\rm 2},\\
    Chao Li\textsuperscript{\rm 3},
    Yunbo Cao\textsuperscript{\rm 2}, and
    Qingyu Zhou\textsuperscript{\rm 2}\footnote{Corresponding author.}
}
\affiliations {
    \textsuperscript{\rm 1} South China University of Technology,
    \textsuperscript{\rm 2} Tencent Cloud Xiaowei,
    \textsuperscript{\rm 3} Xiaomi Group, \\
    \texttt{selizhaoliu@mail.scut.edu.cn, qingyuzhou@tencent.com}
}

\usepackage{bibentry}

\usepackage{pdfpages} 

\newcommand{\chcsd}{Handwritten CCSE\xspace}
\newcommand{\chcsdshort}{CCSE-HW\xspace}
\newcommand{\ckcsd}{Kaiti CCSE\xspace}
\newcommand{\ckcsdshort}{CCSE-Kai\xspace}
\newcommand{\bname}{Chinese Character Stroke Extraction\xspace}
\newcommand{\bnameshort}{CCSE\xspace}
\newcommand{\combineshort}{CCSE-Kai\&HW}

\definecolor{currentStrokeColor}{RGB}{187,68,68}
\def\currentStroke{\textcolor{currentStrokeColor}}
\def\ignore{\textcolor{lightgray}}
\def\imporve{\textcolor{green}}

\def\alec{\textcolor{black}}
\def\lkyy{\textcolor{black}}

\def\lec{\textcolor{black}}
\def\lky{\textcolor{black}}
\def\hsx{\textcolor{black}}

\usepackage{xcolor}

\usepackage{graphicx} 

\usepackage{url}

\begin{document}

\maketitle

\begin{abstract}
Stroke is the basic element of Chinese character and stroke extraction has been an important and long-standing endeavor. Existing stroke extraction methods are often handcrafted and highly depend on domain expertise due to the limited training data. Moreover, there are no standardized benchmarks to provide a fair comparison between different stroke extraction methods, which, we believe, is a major impediment to the development of Chinese character stroke understanding and related tasks. In this work, we present the first public available \bname (\bnameshort) benchmark, with two new large-scale datasets: \ckcsd (\ckcsdshort) and \chcsd (\chcsdshort). With the large-scale datasets, we hope to leverage the representation power of deep models such as CNNs to solve the stroke extraction task, which, however, remains an open question. To this end, we turn the stroke extraction problem into a stroke instance segmentation problem. \alec{Using the proposed datasets to train a stroke instance segmentation model, we surpass previous methods by a large margin. Moreover, the models trained with the proposed datasets benefit the downstream font generation and handwritten aesthetic assessment tasks. We hope these benchmark results can facilitate further research. The source code and datasets are publicly available at: \textcolor{blue}{\texttt{https://github.com/lizhaoliu-Lec/CCSE}}.}
\end{abstract}

\section{Introduction}

\begin{figure}[t]
    \centering
    \includegraphics[width=0.90\linewidth]{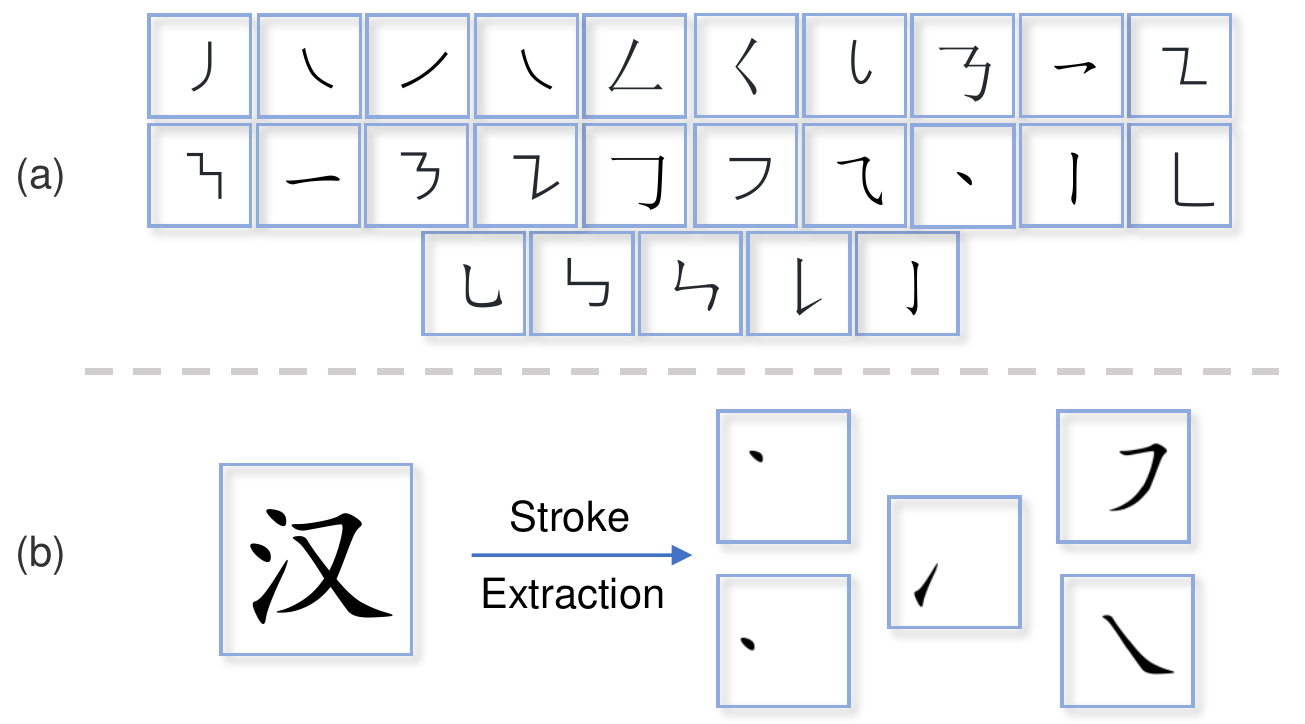}    
    \caption{(a) Illustration of 25 kinds of Chinese character strokes \lec{considered in this paper,} which serve as the building rock of Chinese characters. (b) Illustration of the Chinese character stroke extraction task. Given a Chinese character, the stroke extraction task requires the model to decompose the character into individual strokes.}
    \label{fig:task_definition}
\end{figure}

Stroke is the basic element of Chinese character and \lec{stroke} extraction has been an important and long-standing endeavor~\cite{lee1998chinese}. \lec{Given an image of a Chinese character, stroke extraction aims to decompose it into individual strokes (see Figure~\ref{fig:task_definition}).} It serves as a bedrock for many Chinese character-related applications such as handwritten synthesis~\cite{liu2021fontrl}, font generation~\cite{jiang2019scfont,zeng2021strokegan,xie2021dg}, character style transfer~\cite{huang2020rd}, handwritten aesthetic evaluation~\cite{xu2007intelligent,sun2015aesthetic}, \etc~Recently, it has been shown that explicitly incorporating the stroke information boosts the performance of Chinese character-related tasks~\cite{gao2020gan,huang2020rd,zeng2021strokegan}.
Though various tasks that leverage the stroke information has gained a large amount of attention from the community and made substantial progress by applying the state-of-the-art deep models, the understanding of the Chinese character stroke alone has fallen behind.

\alec{Generally, there are two lines of works}: \alec{stroke extraction from skeleton images~\cite{fan2000run,liu2001model,liu2006geometrical,su2009stroke,zeng2010cascade} and from original images~\cite{lee1998chinese,yu2012stroke}. For skeleton-based methods, the thinning algorithm~\cite{arcelli1985width} is often used as a preprocessing step, which introduces stroke distortion and the loss of short strokes. Stroke extraction from the original image is thereby proposed to address these issues. This kind of approach typically enjoys rich information such as stroke width and curvature, obtaining good performance. The latest research~\cite{xu2016decomposition} proposes to combine merits from both worlds by finding the cross points on the skeleton and combining stroke segments on original images. However, due to the lack of a large-scale dataset to develop learning-based methods, most previous approaches are rule-based and require in-depth expertise during algorithm design. Thus, they \textit{inherently} suffer from the following limitations: \textbf{First}, to decompose the character into stroke segments, handcrafted rules are required to find the partition points, which inevitably contain fork points due to the complex character structure. \textbf{Second}, these methods are typically tailored to the regular and highly structural standard fonts and may not perform well on handwritten characters due to the large intra-class variance of strokes caused by different handwriting habits. \textbf{Last}, they aim to optimize the stroke extraction task \textit{only} and may not produce transferable features to benefit downstream tasks.
}

\lec{Moreover, there are no standardized benchmarks to provide a fair comparison between different stroke extraction methods, which is of great importance to guide and facilitate further research. And the lack of publicly available datasets leads to inconsistent evaluation protocols. Specifically,~\cite{cao2000model,qiguang2004algorithm,xu2016decomposition} consider accuracy as the main evaluation metric for the stroke extraction task, which does not consider the spatial location of the extracted stroke, thereby, can not comprehensively measure the performance of stroke extraction algorithm.~\cite{chen2016benchmark, chen2017automatic} leverage Hamming distance and cut discrepancy to measure the consistency of stroke interiors and the similarity of stroke boundaries, respectively. They require the extracted strokes and the ground truth strokes to be strictly aligned by spatial location and categories, which is hard to evaluate the missed and false extraction. Thus, how to effectively evaluate the stroke extraction algorithm with reasonable protocol remains an unsolved question.
}

\lec{To facilitate stroke extraction research, we present a \bname (\bnameshort) benchmark, with two new large-scale datasets and evaluation methods. As the foundation of the \bnameshort benchmark, the datasets have two requirements: \ie~character-level diversity and stroke-level diversity. Specifically, the datasets should cover as many Chinese characters to represent the structure between strokes, whose relationship can be very complex (see the left of Figure~\ref{fig:dataset_samples}). Moreover, since humans with different writing habits will produce very different appearances even for the same stroke (see the right of Figure~\ref{fig:dataset_samples}), the datasets should cover this kind of diversity for models to achieve effective extraction. To this end, we harvested a large set of \texttt{Kai Ti} (\lky{a kind of Chinese font}) Chinese character images and handwritten Chinese character images to achieve character-level diversity and stroke-level diversity, respectively.}

With the \lec{large-scale} datasets, \lec{we hope to leverage the representation power of deep models such as CNNs to solve the stroke extraction task, which, however, remains an open question.}
\lec{To this end,} we turn the stroke extraction problem into the stroke instance segmentation problem. This change of view \lec{not only} allows us to take advantage of the state-of-the-art instance segmentation models \lec{but also the well-defined evaluation metrics (\ie~box AP and mask AP).}
We perform experiments with state-of-the-art instance segmentation models to produce benchmark results that facilitate further research. \lec{Compared to previous methods of stroke extraction, our approach does not require reference images and in-depth domain expertise. Moreover, the deep models trained on our dataset are able to produce transferable features that \alec{consistently} benefit the downstream tasks.}

We summarize our contributions as follows:
\begin{itemize}
    \item \lec{We propose the first benchmark containing two high-quality large-scale datasets that satisfy the requirements of the character-level and stroke-level diversities for building promising stroke extraction models.}
    \item \lec{We cast the stroke extraction problem into the stroke instance segmentation problem. \lky{In this way, we} build deep stroke extraction models that \alec{scale to scenarios with highly-diverse characters and stroke variance} \alec{while producing transferable features to benefit} downstream tasks.}
    \item \lec{By leveraging the state-of-the-art instance segmentation models and well-defined evaluation metrics, we build standardized benchmarks to facilitate further research.}
\end{itemize}

\section{Related Work}

\subsection{Stroke Extraction} 
\hsx{Stroke extraction aims to extract strokes from handwritten image~\cite{lee1998chinese}, \lec{which is very difficult to solve due to} the \alec{complex character structure}~\cite{cao2000model} and the large intra-class variances~\cite{xu2016decomposition}. \lec{Existing methods mainly follow stroke extraction \alec{from skeletonized character or from original character} paradigms.} For the first kind of approach, \alec{efforts have been put into exploring the relations between strokes by resolving the fork points issues~\cite{fan2000run}, applying affine transformation to strokes~\cite{liu2006geometrical}, detecting ambiguous zone~\cite{su2009stroke} and using additional reference image~\cite{zeng2010cascade}.} \alec{However, these approaches are limited by the thinning step that introduces stroke distortion and the loss of short strokes. Therefore, stroke extraction from the original image is proposed to conquer this limitation. These approaches focus on leveraging the rich information in characters such as stroke width and curvature by combining multiple contour information in strokes~\cite{lee1998chinese}, exploring pixel-stroke relationships~\cite{cao2000model}, detecting strokes in multiple directions~\cite{su2004decomposing} and using corner points~\cite{yu2012stroke}.} \alec{The latest approach~\cite{xu2016decomposition} considers the advantages from both worlds to further improve the performance. Nonetheless, \lec{these} methods typically use handcrafted rules to improve the stroke extraction task only during algorithm design. Therefore, they inherently suffer from extracting strokes from complex characters and with highly irregular shape. Moreover, they can not be trivially employed for downstream tasks such as font generation, limiting their further application.}
}

\subsection{Instance Segmentation}
\hsx{The goal of instance segmentation is to segment every instance (countable objects) in an image by assigning it with pixel-wise class label. 
Existing approaches can be broadly divided into two categories: {two-stage}~\cite{he2017mask,hsieh2021droploss} and {one-stage}~\cite{Bolya_2019_ICCV}.
\lky{Two-stage methods consist of instance detection and segmentation steps.} In Mask R-CNN~\cite{he2017mask}, \lec{one of the most important milestones in computer vision,} the segmentation head is applied to the detected instances from the Faster R-CNN~\cite{ren2015faster} detector to acquire the instance-wise segmentation mask. 
Approaches based on Mask R-CNN typically demand dense prior proposals or anchors to obtain decent results, leading to complicated label assignment and post-processing steps. 
To tackle this issue, one-stage methods such as YOLACT~\cite{Bolya_2019_ICCV} produce instance masks by linearly combining the prototypes with the mask coefficients and \lec{do not} depend on \lec{pre-detection step.}
In this paper, we \alec{benefit from the rapid development of instance segmentation algorithms and} focus on applying the instance segmentation models to tackle the stroke extraction task, thus we mainly consider the well-studied two-stage methods such as Mask R-CNN as our baselines.
}





\begin{figure}[thb]
    \centering
    \includegraphics[width=\linewidth]{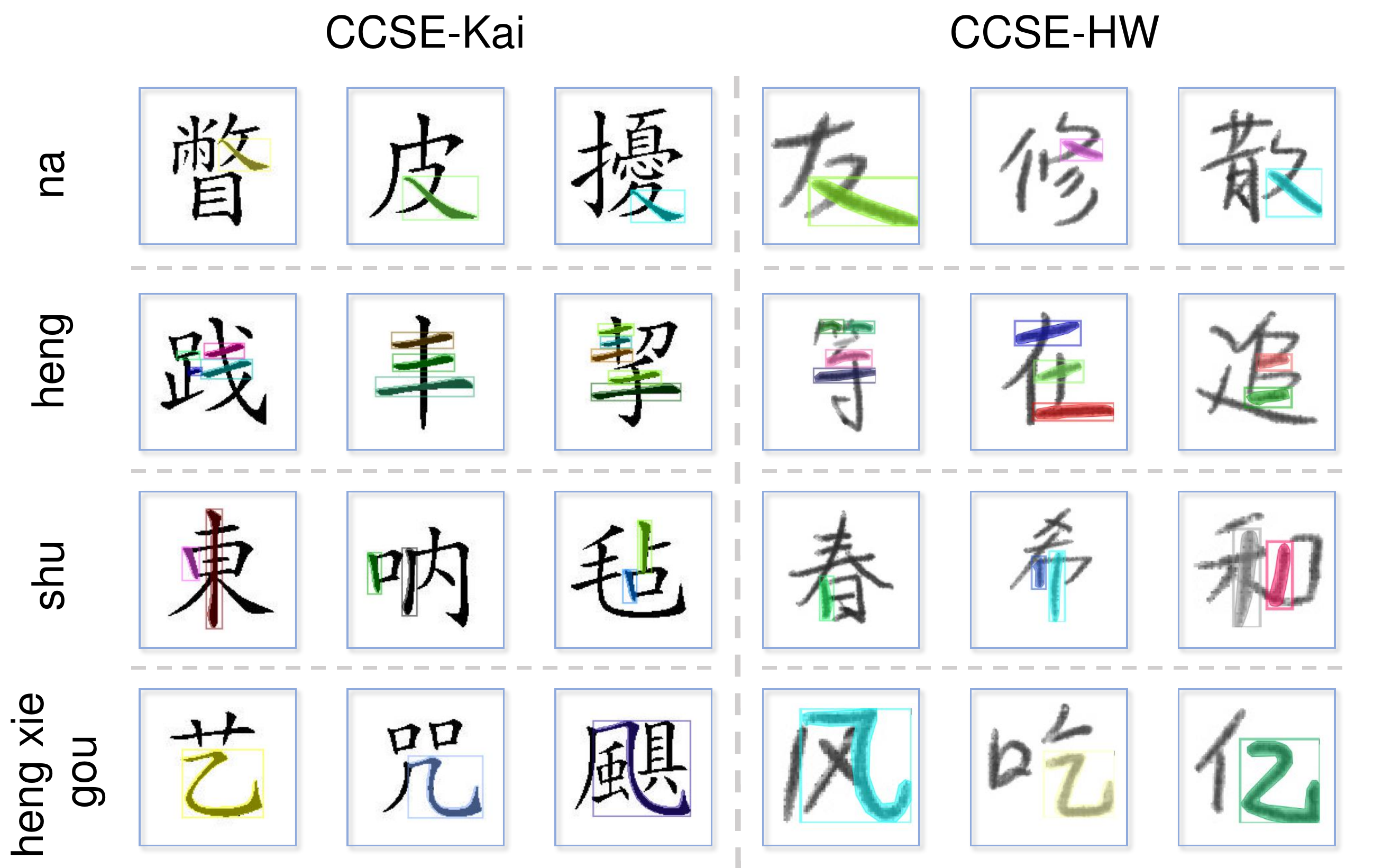}    
    \caption{From left to right, samples of annotated Chinese characters in \ckcsdshort dataset and \chcsdshort dataset.}
    \label{fig:dataset_samples}
\end{figure}

\section{Proposed Datasets}

\subsection{Image Collection and Annotation}
\lec{To achieve promising stroke extraction performance, we harvest a large number of samples that cover the complex structures of Chinese characters and different styles of stroke, which are character-level and stroke-level diversity, respectively. Since the frequently used Chinese characters are restricted to a small range, there may not have enough handwritten characters with complex stroke structures. Thus, we collect the frequently used standard font (\eg~\texttt{Kai Ti}) to meet the character-level diversity requirement. Then, to satisfy the stroke-level diversity, we gather handwritten Chinese character images from different writers. We detail the process of collection and annotation below.}

\subsubsection{\texttt{Kai Ti} Image Collection and Annotation}

\begin{figure}[thb]
    \centering
    \includegraphics[width=\linewidth]{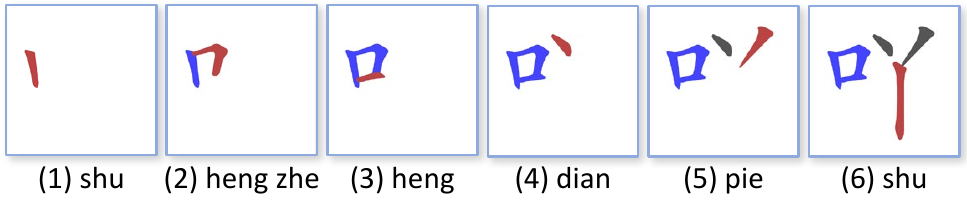}    
    \caption{Illustration of the \texttt{Kai Ti} image collection process. We use the open source character rendering library cnchar to generate the images of a Chinese character in a stroke incremental manner. The character \texttt{ya} is written stroke by stroke with the stroke highlighted by \currentStroke{light brown} in the image and the stroke class denoted underneath.}
    \label{fig:KaitiImage}
\end{figure}

\lec{Labeling every stroke in an image is time-consuming and labor-intensive. Since \texttt{Kai Ti} is a standard Chinese font commonly used in daily life,} our first thought is to collect an annotation-free \texttt{Kai Ti} dataset by retrieving the spatial information from its font design database. However, the coordinates of each stoke are not preserved \lec{during} the font design process. Thus, we \lec{browse the web resources} extensively and discover an open source project Make Me A Hanzi\footnote{\url{https://github.com/skishore/makemeahanzi}}, which has constructed a stroke database for \texttt{Kai Ti}. Then, this project is further \lec{evolved} by cnchar\footnote{\url{https://github.com/theajack/cnchar}}, which provides more user-friendly interfaces to access the \texttt{Kai Ti} image stroke-by-stroke. \lec{As shown in Figure~\ref{fig:KaitiImage}, the results from cnchar} have a clear stroke-wise mark with \currentStroke{light brown} denoting the spatial mask and category of the current stroke. \lec{Regarding the stroke category, the database of cnchar contains the most frequently used 25 categories (see Figure~\ref{fig:task_definition} (a) for details).}

\begin{figure}[thb]
    \centering
    \includegraphics[width=\linewidth]{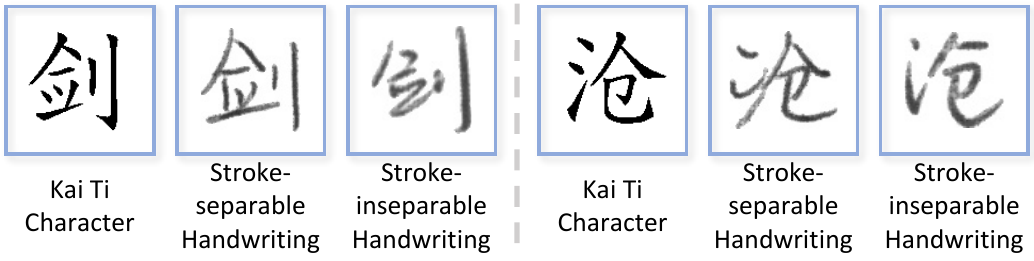}
    \caption{Comparison between the stroke-separable and stroke-inseparable handwriting. The corresponding \texttt{Kai Ti} characters are put on the left for reference.}
    \label{fig:handwrittenStrokeSeparable}
\end{figure}

\lec{With the assistance of cnchar, we harvest stroke-wise images from \textbf{9,523} unique \texttt{Kai Ti} Chinese characters. Then, we use OpenCV\footnote{\url{https://opencv.org/}} to produce the bounding box and mask annotation from the \currentStroke{light brown} area, resulting in our \ckcsd (\ckcsdshort) dataset. The visualization results of \ckcsdshort are depicted on the left of Figure~\ref{fig:dataset_samples}. We can see that \ckcsdshort provides samples with complex stroke structures. There are more than \textbf{1M} stroke instances in \ckcsdshort and the detailed statistics will be elaborated later. The merits of our \ckcsdshort are as follows: 1) We discover an automated method to effectively produce a stroke instance dataset without extensive human labor. 2) \ckcsdshort satisfies the character-level diversity by covering most of the Chinese characters despite the usage frequency. However, its shortcoming is obvious: lack of stroke-level diversity since the stroke in the standard font library is relatively fixed. In this sense, the model trained with \ckcsdshort may not deliver satisfactory results in some application scenarios, where extracting strokes from handwritten Chinese is desired.} 

\subsubsection{Handwritten Image Collection and Annotation}
\lec{Since \ckcsdshort only meets 
character-level diversity, we target at improving the stroke-level diversity of our dataset by leveraging the handwritten character with various styles.} To \lec{this end}, we further harvest handwritten Chinese characters and label them in a stroke instance manner. Specifically, we leverage the CASIA Offline Chinese Handwriting Databases\footnote{\url{http://www.nlpr.ia.ac.cn/databases/handwriting/Home.html}}, which has 7,185 kinds of Chinese characters written repeatedly by about 300 humans, resulting in nearly 3M handwritten Chinese images.

\begin{figure*}
    \centering
    \begin{subfigure}[c]{0.30\textwidth} 
        \centering
        \includegraphics[width=\textwidth]{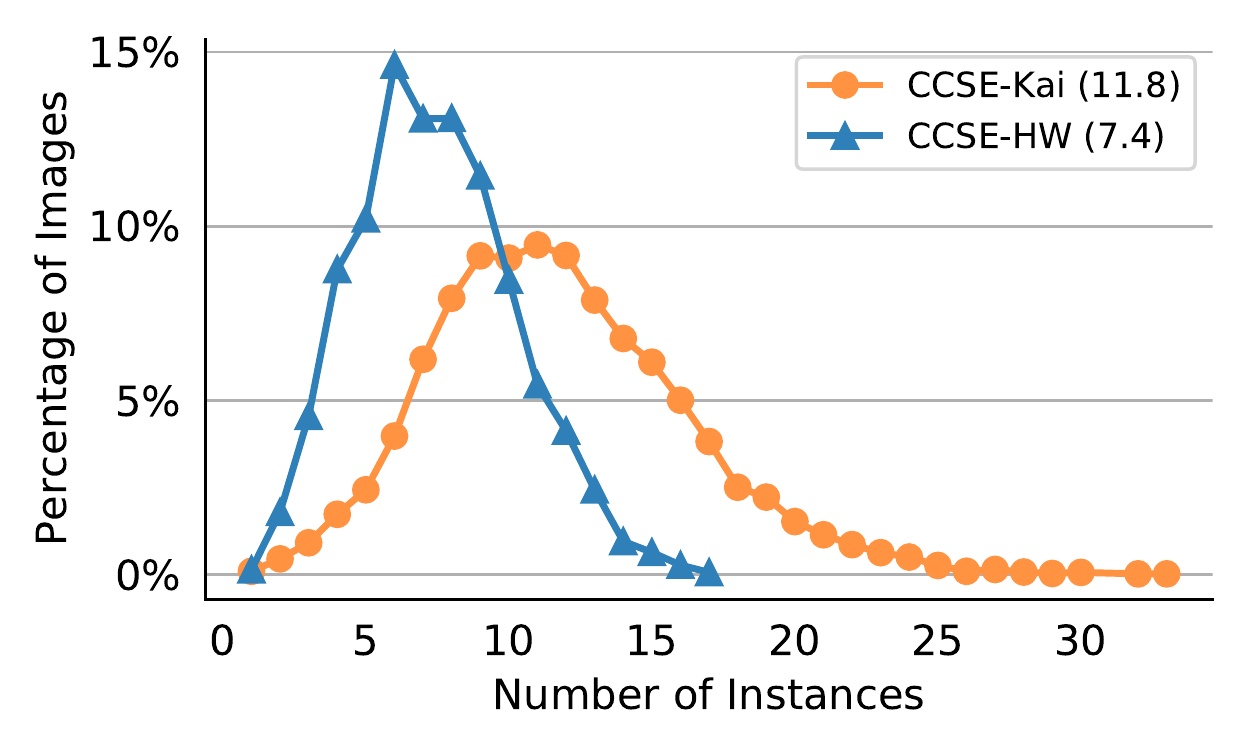}
        \caption{Number of instances in an image.}
        \label{fig:instances_per_image}
    \end{subfigure}
    \begin{subfigure}[c]{0.302\textwidth} 
        \centering
        \includegraphics[width=\textwidth]{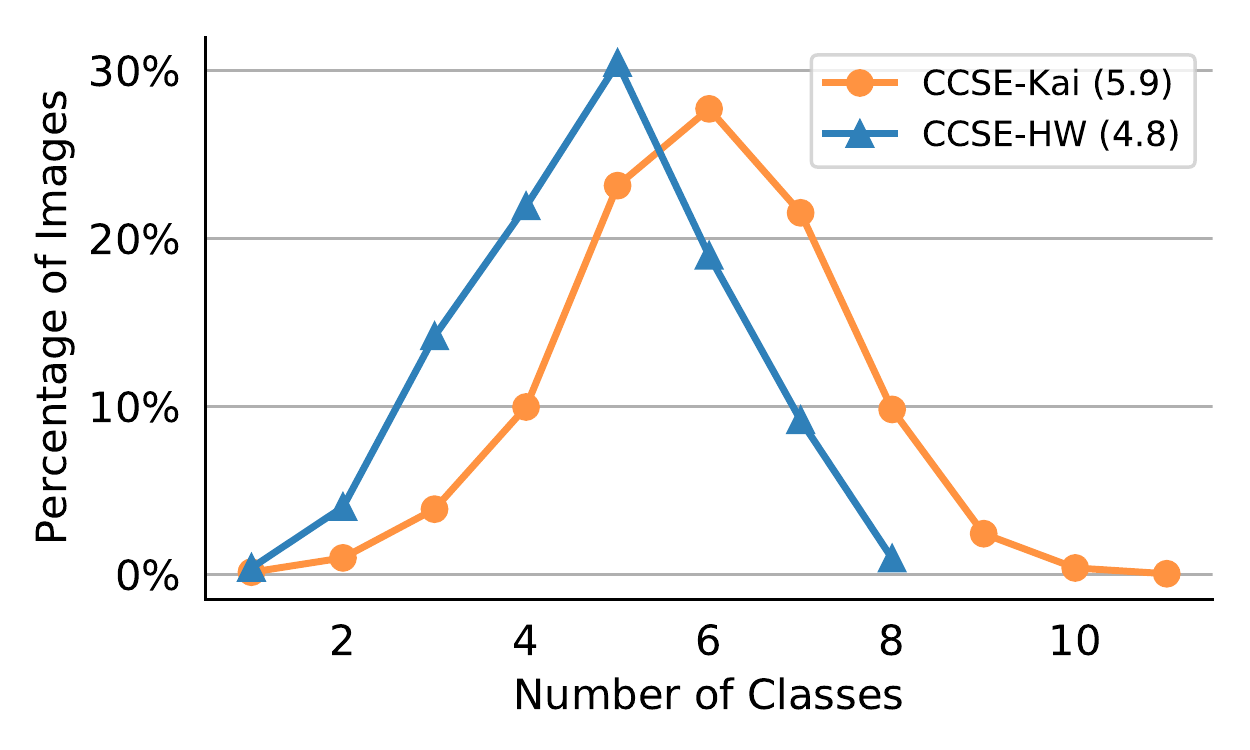}
        \caption{Number of categories in an image.}        
        \label{fig:classes_per_image}
    \end{subfigure}
    \begin{subfigure}[c]{0.302\textwidth} 
        \centering
        \includegraphics[width=\textwidth]{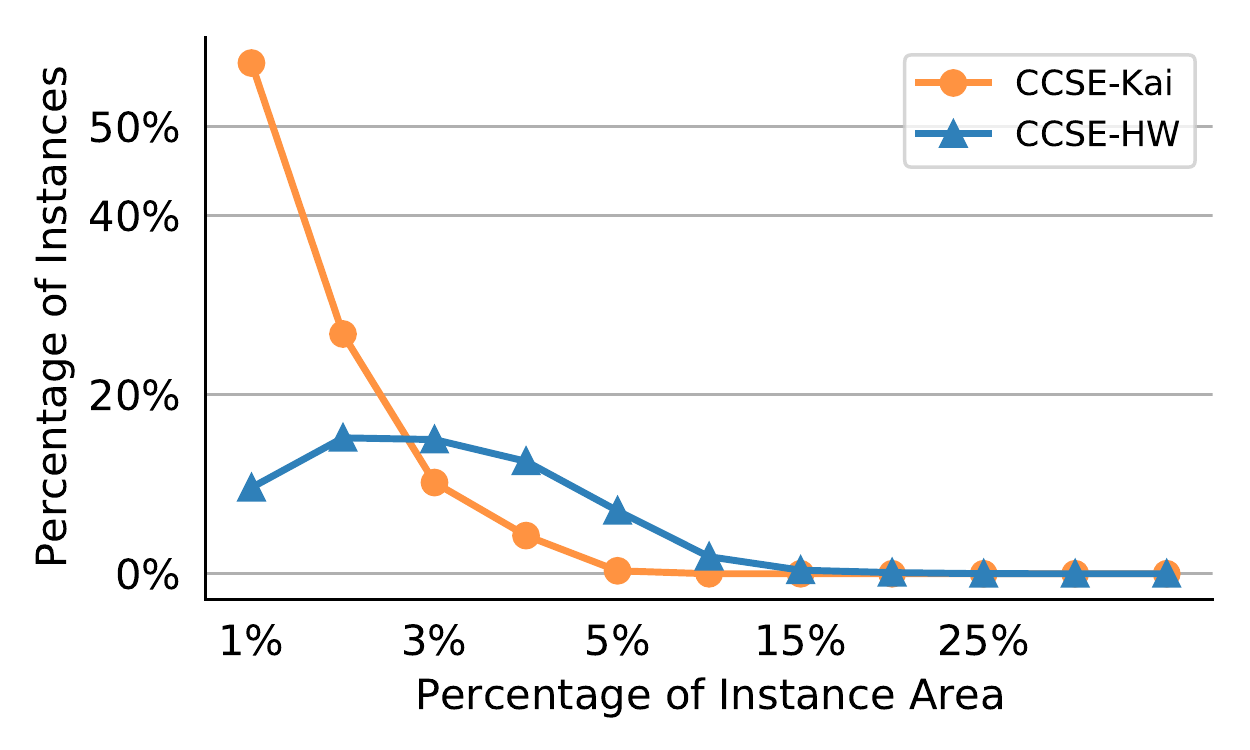}
        \caption{\lec{Instance area \vs~Number of Instances.}}            
        \label{fig:area_percentage}
    \end{subfigure}
    \caption{The statistics of stroke instances in \chcsdshort and \ckcsdshort datasets.}
    \label{fig:detailed_analysis}
\end{figure*}

However, as shown in Figure~\ref{fig:handwrittenStrokeSeparable}, some human writers draw a character that is not stroke-separable, \lec{which can not be trivially handled in stroke extraction task.} To \lec{tackle} this issue, we \lec{sub-sample the data} that is stroke-separable from CASIA. Moreover, considering that human annotation is \lec{labor-intensive} and time-consuming, we select 10 samples for the \lec{top 300} most frequently used Chinese characters and 8 samples for the next  700 Chinese characters, resulting in about 7,600 \lec{images in total}. Then, we apply extensive human labor to carefully provide annotation for each stroke and \lky{finally} create a \chcsd (\chcsdshort) dataset. \lec{Note that we adopt the stroke categories used in \ckcsdshort during the stroke annotation process.} \lec{The visualization results of \chcsdshort are shown on the right of Figure~\ref{fig:dataset_samples}, from which we can see that strokes of the same category appear very differently in terms of scale, coverage and curvature \etc~\lky{So far}, we overcome the shortcoming of \ckcsdshort by complementing the stroke-level diversity. With both \ckcsdshort and \chcsdshort, we provide datasets with rich character and stroke-level diversity to build our benchmarks effectively and reasonably.}


\subsection{Dataset Statistics}
In this section, we analyze the properties of the proposed \ckcsdshort and \chcsdshort datasets. \lec{We first compare our datasets to existing datasets with respect to the amount and annotation type. Then, we analyze the proposed datasets and intrinsic difficulties that occurred in our datasets.}

\begin{table}[htb]
    \centering
    \resizebox{\linewidth}{!}{
    \begin{tabular}{lccrr}
    \toprule
    Dataset & Pub. Ava. & Annotation Type & \#Images & \#Strokes \\
    \midrule
    \cite{cao2000model} & $\times$ & category & 111 & 849 \\
    \cite{xun2015stroke} & $\times$ & category & 518 & \na \\
    \cite{xu2016decomposition} & $\times$ & category & 1,500 & \na \\
    \cite{chen2016benchmark} & $\times$ & category & 2,556 & \na \\
    \ckcsdshort (Ours) & $\checkmark$ & instance mask & 9,523 & 112,024 \\
    \chcsdshort (Ours) & $\checkmark$ & instance mask & 7,628 & 56,722 \\
    \combineshort (Ours) & $\checkmark$ & instance mask & \textbf{17,151} & \textbf{168,746} \\
    \bottomrule
    \end{tabular}
    }
    \caption{Comparison between different Chinese character stroke datasets. We propose the largest publicly available Chinese stroke datasets with instance mask annotation to date. Pub. Ava. is short for Publicly Available.}
    \label{tab:compare_dataset}
\end{table}

\subsubsection{Comparison to Existing Datasets}
\lec{We analyze the size of the proposed datasets in comparison to several commonly used datasets~\cite{cao2000model,xun2015stroke,xu2016decomposition,chen2016benchmark} for Chinese stroke extraction. The summary is shown in Table~\ref{tab:compare_dataset}. We have about $4\times$ amount of images compared to the previous largest one (\eg~9,523 \vs~2,556). Notably, different from existing datasets that only provide category level labels, we provide an instance level mask for each stroke, which contains detailed spatial as well as shape information. Most importantly, we are the first one \lky{to} provide publicly available datasets for stroke extraction, facilitating fair comparisons of stroke extraction and downstream tasks.}

\begin{figure}
    \centering
    \includegraphics[width=\linewidth]{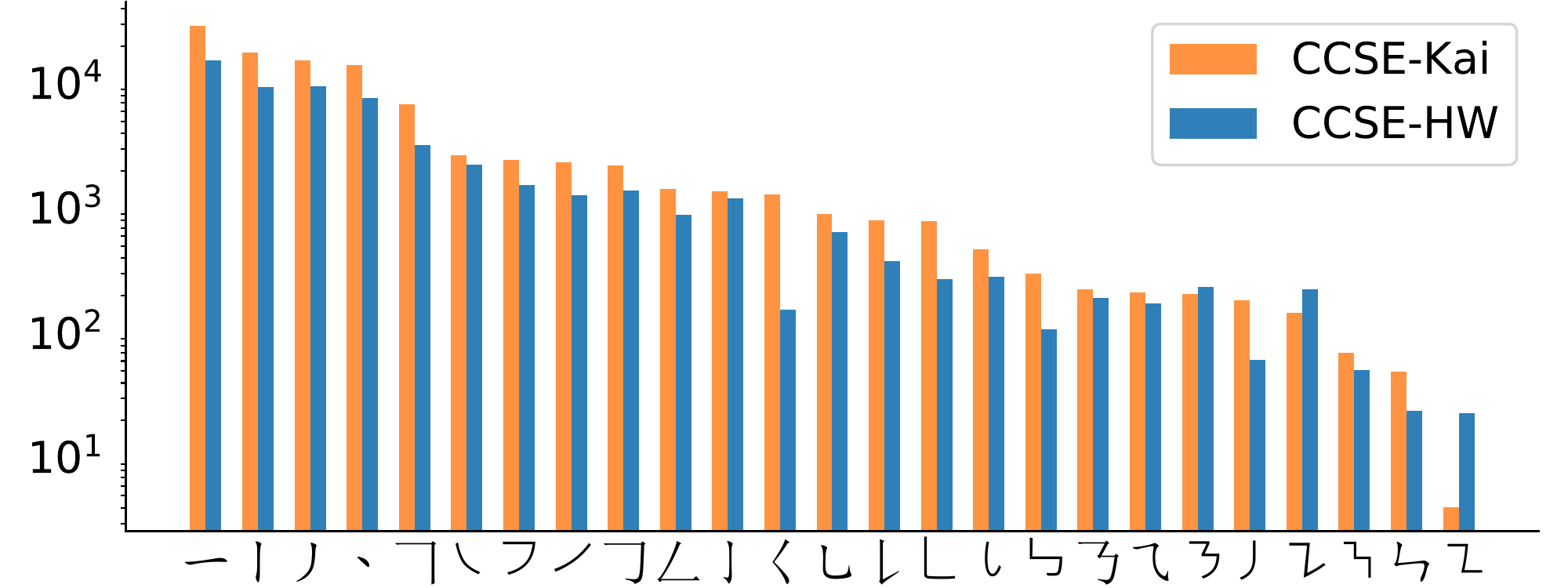}
    \caption{The number of annotated instances per category in \chcsdshort and \ckcsdshort.}
    \label{fig:instances_datasets}
\end{figure}

\begin{figure}
    \centering
    \includegraphics[width=\linewidth]{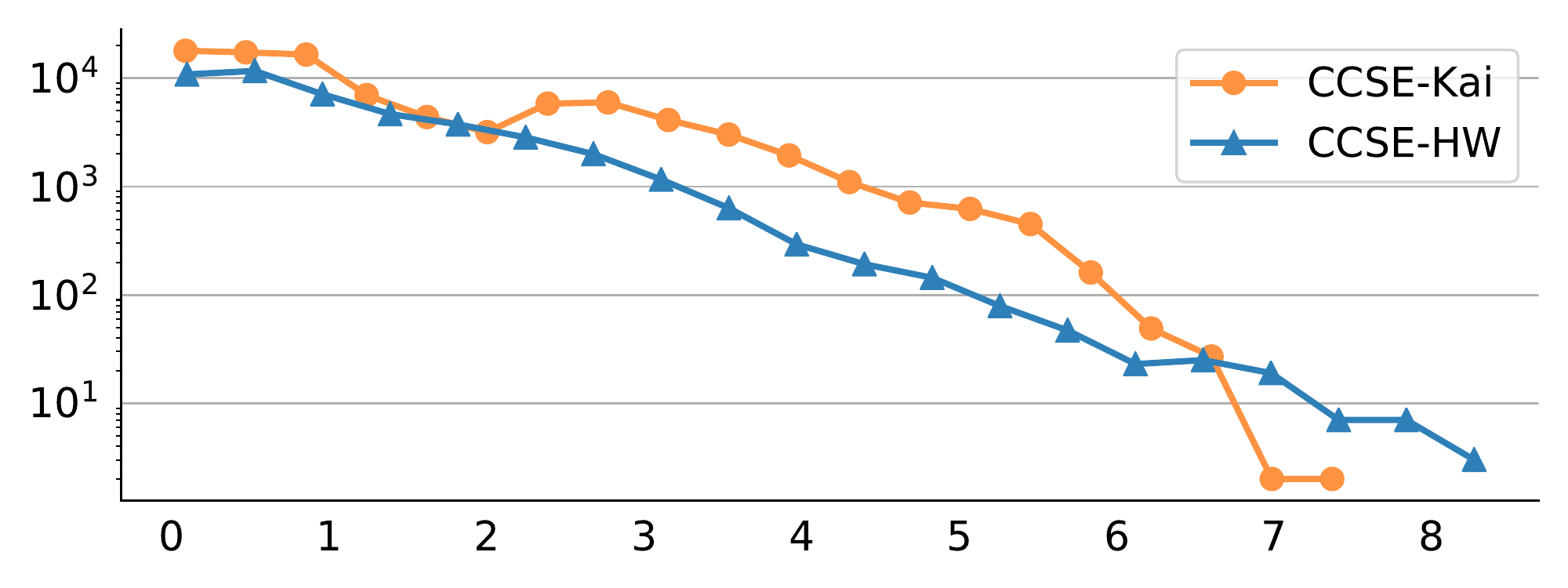}
        \caption{The cumulative distribution histogram of bounding box scale ratios in our two proposed datasets.}
    \label{fig:anchor_scale}
\end{figure}

\begin{figure*}[!htb]
    \centering
    \includegraphics[width=\linewidth]{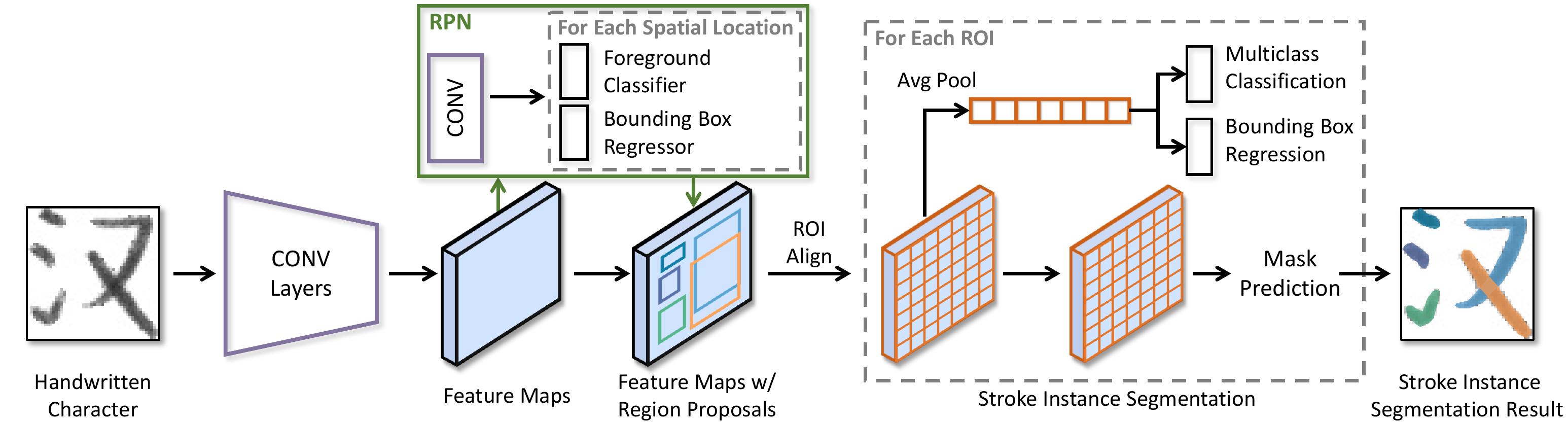}
    \caption{Overview of the stroke instance segmentation model for stroke extraction task (using Mask R-CNN for illustration).}
    \label{fig:sparsercnn}
\end{figure*}

\subsubsection{Analysis on \ckcsdshort and \chcsdshort}
\lky{
We \lec{mainly} perform quantitative analyses on \lec{our datasets in terms of instance level and category level. The results are shown in Figure~\ref{fig:detailed_analysis}. From Figure~\ref{fig:instances_per_image} and Figure~\ref{fig:classes_per_image}, we observe that \ckcsdshort provides more strokes in one image in averaged as we expected since complex stroke structures typically introduce more strokes and categories in one character. This shows that \ckcsdshort indeed improves the character-level diversity for our benchmark datasets. Moreover, as depicted in Figure~\ref{fig:area_percentage}, we find that \chcsdshort covers a wider range in an image, which suggests that the handwritten character is able to improve the stroke-level diversity by including strokes with various scales. These results verify that our datasets fulfill the diversity requirements for achieving promising stroke extraction performance.} 
}

\lec{We then reveal the intrinsic difficulties of our datasets by analyzing the number of strokes per category and the scale statistics of our bounding box, where the results are shown in Figure~\ref{fig:instances_datasets} and Figure~\ref{fig:anchor_scale}, respectively. From Figure~\ref{fig:instances_datasets}, we observe that the stroke extraction task faces a severe class imbalance problem, which may results in impeded performance for classifying strokes with few data points. Moreover, we also find out from Figure~\ref{fig:anchor_scale} that: 1) strokes are often in a strip shape, which is a major difference from the common object detection. 2) the shape of stroke also occurs a class imbalance problem, making it difficult to locate the stroke with a very strip shape. Solving these difficulties is out of the scope of this paper and we leave them to our future works.}

\begin{table}
    \centering
    \resizebox{\linewidth}{!}{
    \begin{tabular}{lc|ccc|ccc}
    \toprule
    Model & $\mD$ & $\text{AP}_{50}^{\text{box}}$ & $\text{AP}_{75}^{\text{box}}$ & $\text{AP}^{\text{box}}$ & $\text{AP}_{50}^{\text{mask}}$ & $\text{AP}_{75}^{\text{mask}}$ & $\text{AP}^{\text{mask}}$ \\
    \midrule
    Mask R-CNN & $K$ & {93.15} & {82.70} & {78.73} & \textbf{57.68} & \textbf{52.01} & \textbf{44.89} \\
    Cascade R-CNN & $K$ & \textbf{94.21} & \textbf{91.91} & \textbf{80.32} & {55.97} & {50.40} & {43.35} \\
    \midrule
    Mask R-CNN & $H$ & \textbf{90.73} & {83.03} & {72.09} & \textbf{92.29} & {81.26} & {68.27} \\
    Cascade R-CNN & $H$ & {89.70} & \textbf{83.27} & \textbf{74.76} & {90.71} & \textbf{83.15} & \textbf{68.71} \\
    \bottomrule
    \end{tabular}
    }
    \fontsize{8}{9}\selectfont
    \caption{Experiment results of Mask R-CNN and Cascade Mask R-CNN. $\mD$ is the abbreviation of Dataset. 
     $K$ and $H$ are short for \ckcsdshort and \chcsdshort, respectively.
    }
    \label{tab:main_results}
\end{table}

\section{Algorithmic Analysis}

\noindent{\textbf{Baseline}. \lec{To build stroke detection baselines\footnote{\alec{Results are in the supplementary.}}}, we \lec{consider} widely used detectors Faster R-CNN~\cite{ren2015faster}, Cascade R-CNN~\cite{CascadeRCNN_2018_CVPR} and FCOS~\cite{tian2019fcos}. \lec{For constructing stroke instance segmentation benchmark results, we employ Mask R-CNN~\cite{he2017mask} and its cascade version~\cite{CascadeRCNN_2018_CVPR}. The overview of stroke instance segmentation workflow is depicted in Figure~\ref{fig:sparsercnn}.}} \alec{For simplicity, we use $K$ and $H$ to denote \ckcsdshort and \chcsdshort datasets, respectively.}

\noindent{\textbf{Implementation details}. Our implementation is based on detectron2~\cite{wu2019detectron2} framework. \lec{Since the training cost for our datasets is low due to low image resolution, we apply the 3$\times$ training schedule by default. All experiments are performed on a single Titan XP GPU. The minimum training image sizes are randomly selected from $\{112, 120\}$ for each iteration. For bounding box regression, we use the generalized IoU loss by default. As for other hyper-parameters and module choices, we follow the default settings in detectron2. Mask R-CNN is used as our default stroke instance segmentation model. As for train/val/test partition, we randomly spilt both \ckcsdshort and \chcsdshort with ratio 9:1:1.}}

\subsection{Stroke Instance Segmentation}
\subsubsection{Main Results}
\begin{figure}[!t]
    \centering
    \includegraphics[width=1\linewidth]{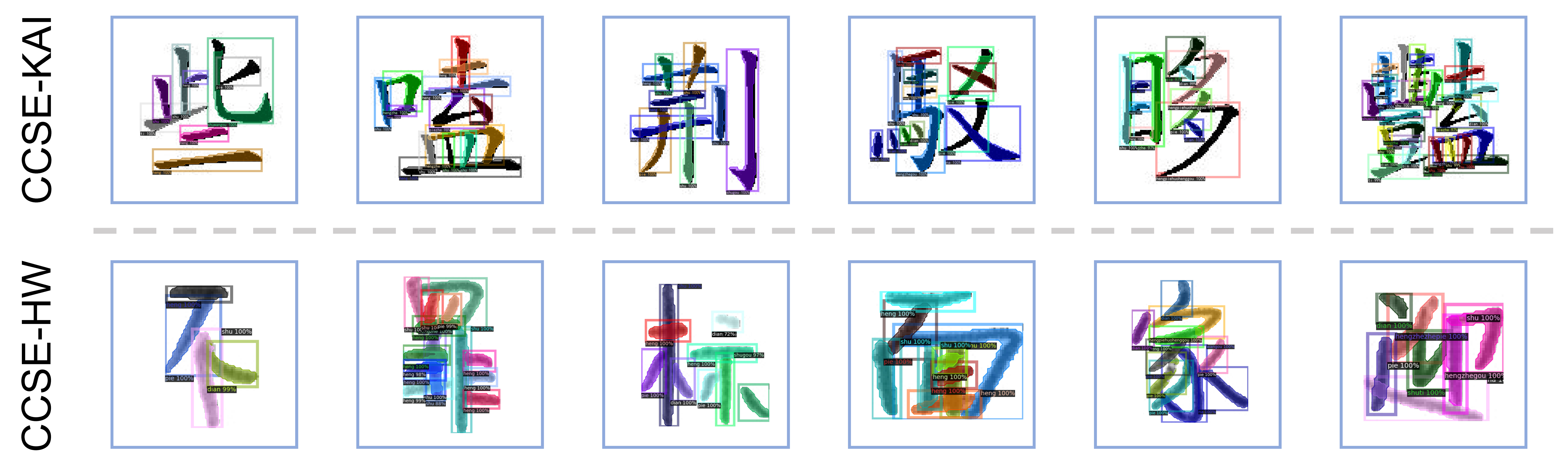}
    \caption{\bnameshort results used Mask R-CNN on \ckcsdshort and \chcsdshort. Best view with zoom in.}
    \label{fig:prediction_visualization}
\end{figure}

\lec{In this section, we present the results of stroke instance segmentation. The quantitative results are in Table~\ref{tab:main_results}. We also provide the qualitative results in Figure~\ref{fig:prediction_visualization}. As can be seen in Table~\ref{tab:main_results}, we achieve promising results for stroke instance segmentation for both \ckcsdshort and \chcsdshort. 
The AP$^\text{mask}$ is low for \ckcsdshort. \alec{We attribute it to the complex characters with many strokes that highly overlapped with each other in \ckcsdshort. It may} be further improved by tailoring the model with complex character structure prior. Notably, as depicted in Figure~\ref{fig:prediction_visualization}, we are able to produce stroke instance segmentation results with a high confidence score, indicating the effectiveness of our datasets and \alec{applying instance segmentation for stroke extraction}. \alec{Due to the space limit, we put the failure case analysis in the supplementary.}}

\subsubsection{Transferability Results on Standard Fonts}
\alec{One may ask whether the proposed dataset can contain character images in more printing font styles that are also stroke-separable. Simply labeling more frequently used printing font styles will fulfill this goal but also be time-consuming and labor-intensive. Considering the highly similar structure and appearance of commonly used font styles (\eg~\texttt{Kai Ti}, \texttt{Song Ti}, \texttt{Hei Ti}), we thus leverage the model trained by our \ckcsdshort dataset to automatically label character images of other font styles. As shown in Figure~\ref{fig:font_trans}\footnote{\alec{More results are put in the supplementary.}}, minor effort is required to adjust the bounding box and mask to use the labels derived by the model trained by our \ckcsdshort.}

\begin{figure}[hb]
    \centering
    \includegraphics[width=\linewidth]{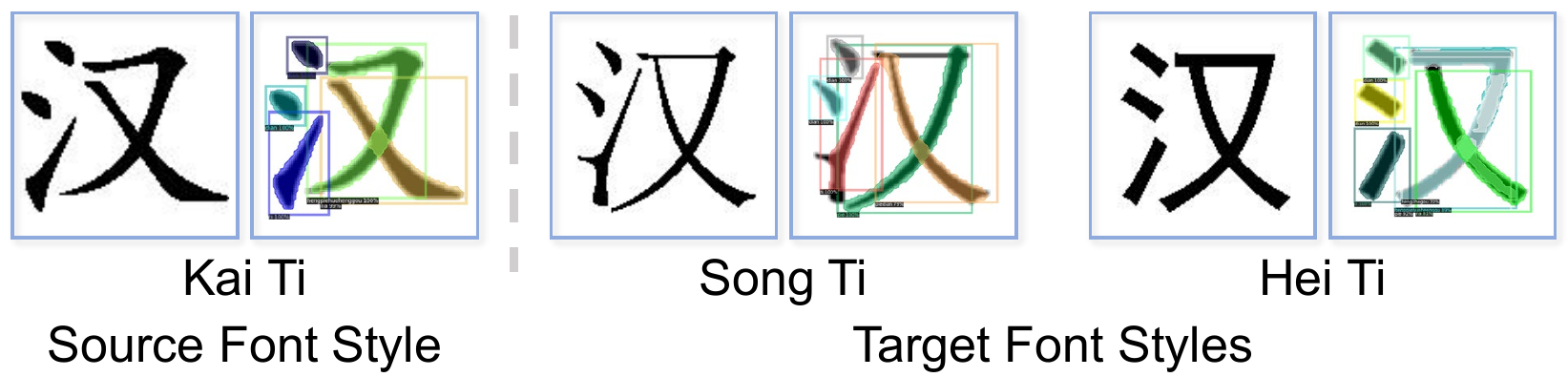}
    \vspace{-0.5cm}
    \caption{Stroke extraction results for \texttt{Song Ti}, \texttt{Hei Ti} styles with the model trained on the \texttt{Kai Ti} dataset.}
    \label{fig:font_trans}
\end{figure}

\subsubsection{Effect of the Background}
\begin{figure}[htb]
    \centering
    \includegraphics[width=\linewidth]{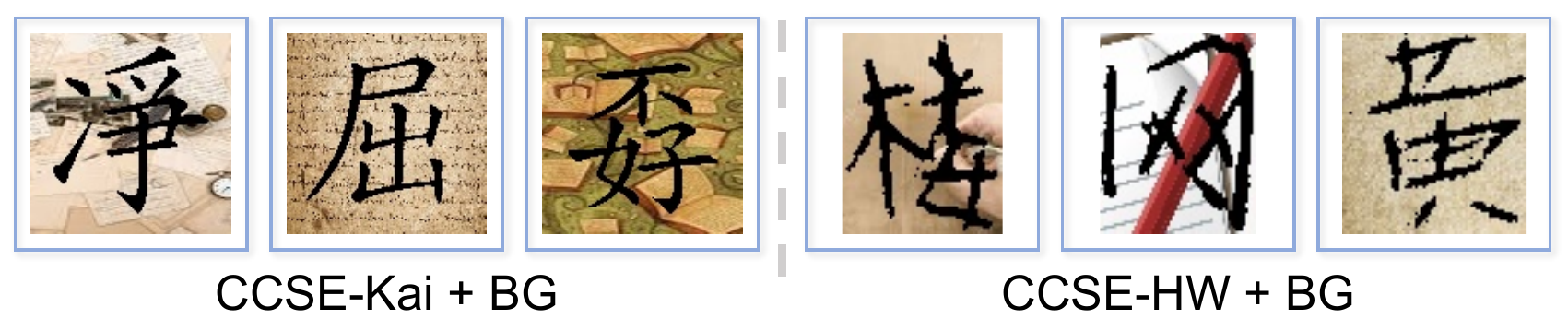}
    \vspace{-0.5cm}
    \caption{From left to right, adding noise background to the samples of \ckcsdshort and \chcsdshort, respectively.}
    \label{fig:dataset_BG}
\end{figure}

\alec{Since the proposed datasets have no background, training a model under this setting may not be suitable for real-life applications with noisy backgrounds. Thus, we conduct experiments to verify and remedy this issue.} \lkyy{As shown in Figure~\ref{fig:dataset_BG}, we add complex backgrounds to character images\footnote{\alec{More results are shown in the supplementary.}} and use them to test the model trained with our original datasets. As shown in Table~\ref{tab:bg}, the performance drops considerably. To compensate for this, we propose to train the model with complex background augmented images, which boosts the performance substantially.
}

\begin{table}[!htb]
    \centering
    \resizebox{\linewidth}{!}{
    \begin{tabular}{ll|cc||ll|cc}
    \toprule
    Train set & Test set & $\text{AP}^{\text{box}}$ & $\text{AP}^{\text{mask}}$ & Train set & Test set & $\text{AP}^{\text{box}}$ & $\text{AP}^{\text{mask}}$ \\
    \midrule
    $K$ & $K$ & {78.73} & {44.89} & $H$ & $H$ & {72.09} & {68.27} \\
    \midrule
    $K$ & $K$ + BG & {72.76} & {36.62} & $H$ & $H$ + BG & {24.40} & {14.91} \\
    $K$ + BG & $K$ + BG & \textbf{76.66} & \textbf{39.83} & $H$ + BG & $H$ + BG & \textbf{61.20} & \textbf{57.06} \\
    \bottomrule
    \end{tabular}}
    \fontsize{8}{9}\selectfont
    \caption{Experiment results of Mask R-CNN on images with a complex background.
    BG denotes adding a complex background to the images in the dataset.
    }
    \label{tab:bg}
\end{table}

\subsubsection{Cross-domain Evaluation}
To evaluate the \lec{robustness} of a trained stroke extraction model, we perform experiments under the cross-domain settings. To be specific, we train the model on the source ($\mS$) training set and evaluate it on the target ($\mT$) test set. \lec{Thus, as shown in Table~\ref{tab:cross_domain_results}, we perform experiments with $(\mS, \mT) \in \{(H, K), (K, H)\}$. The cross-domain evaluation results show that the model is unable to deliver satisfactory performance due to the domain discrepancy caused by unmatched character-level and stroke-level diversities. \alec{Thus, we propose a simple remedy by combing source and target datasets to train the model. In this way, the overall performance is improved compared to using only one dataset. We think there is a more data-efficient way to tackle the domain discrepancy issue such as unsupervised domain adaptation~\cite{ganin2015unsupervised}.}}

\begin{table}[!h]
    \centering
    \resizebox{0.95\linewidth}{!}{
    \begin{tabular}{l|c|ccc|ccc}
    \toprule
    \makecell[c]{$\mS$} & \makecell[c]{$\mT$} & $\text{AP}_{50}^{\text{box}}$ & $\text{AP}_{75}^{\text{box}}$ & $\text{AP}^{\text{box}}$ & $\text{AP}_{50}^{\text{mask}}$ & $\text{AP}_{75}^{\text{mask}}$ & $\text{AP}^{\text{mask}}$ \\
    \midrule
    $K$ & $K$ & {93.15} & {82.70} & {78.73} & {57.68} & {52.01} & {44.89} \\
    $H$ & $K$ & {68.85} & {51.62} & {44.01} & {41.58} & {6.37} & {18.05} \\
    \alec{$H + K$} & $K$ & \textbf{ 94.24 } & \textbf{ 91.71 } & \textbf{ 79.91 } & \textbf{ 59.10 } & \textbf{ 55.92 } & \textbf{ 46.46 } \\
    \midrule
    $H$ & $H$ & {90.73} & {83.03} & {72.09} & \textbf{92.29} & {81.26} & {68.27} \\
    $K$ & $H$ & {29.16} & {6.45} & {11.27} & {4.04} & {0.00} & {0.01} \\
    \alec{$H + K$} & $H$ & \textbf{ 91.52 } & \textbf{ 85.06 } & \textbf{ 72.99 } & { 91.96 } & \textbf{ 83.84 } & \textbf{ 69.56 } \\
    \bottomrule
    \end{tabular}
    }
    \fontsize{8}{9}\selectfont
    \caption{Experiments on different sources and targets. $\mS$ and $\mT$ are the abbreviations of Source and Target respectively.
    }
    \label{tab:cross_domain_results}
\end{table}

\subsection{\alec{Comparison to Previous Approach}}

\begin{table}[!ht]
    \centering
    \resizebox{\linewidth}{!}{
    \begin{tabular}{ll|cccc}
    \toprule
    Method & $\mD$ & Acc. & Prec. & Rec. & F1 \\
    \midrule
    Traditional Approach & $K^{*}_{s}$ & { 35.53 } & { 65.22 } & { 25.94 } & { 34.18 } \\
    Mask R-CNN $_{\text{IoU} \geq 0.9}$ & $K^{*}_{s}$ & \textbf{ 49.52 } & \textbf{ 90.68 } & \textbf{ 66.76 } & \textbf{ 74.78 } \\
    \midrule
    Traditional Approach & $K_{s}$ & { 41.98 } & { 86.36 } & { 42.17 } & { 53.99 } \\
    Mask R-CNN $_{\text{IoU} \geq 0.9}$ & $K_{s}$ & \textbf{ 68.08 } & \textbf{ 90.21 } & \textbf{ 72.55 } & \textbf{ 79.34 } \\
    \ignore{Mask R-CNN $_{\text{IoU} \geq 0.8}$} & \ignore{$K_{s}$} & \ignore{ 90.57 } & \ignore{ 90.15 } & \ignore{ 80.39 } & \ignore{ 84.42 } \\
    \ignore{Mask R-CNN $_{\text{IoU} \geq 0.7}$} & \ignore{$K_{s}$} & \ignore{ 94.97 } & \ignore{ 90.17 } & \ignore{ 81.89 } & \ignore{ 85.27 } \\
    \midrule
    Traditional Approach & $H_s$ & { 36.75 } & { 72.00 } & { 35.60 } & { 45.52 } \\
    Mask R-CNN $_{\text{IoU} \geq 0.9}$ & $H_s$ & \textbf{ 59.54 } & \textbf{ 78.25 } & \textbf{ 56.71 } & \textbf{ 64.78 } \\
    \ignore{Mask R-CNN $_{\text{IoU} \geq 0.8}$} & \ignore{$H_s$} & \ignore{ 82.07 } & \ignore{ 86.33 } & \ignore{ 73.94 } & \ignore{ 79.00 } \\
    \ignore{Mask R-CNN $_{\text{IoU} \geq 0.7}$} & \ignore{$H_s$} & \ignore{ 90.52 } & \ignore{ 90.36 } & \ignore{ 83.78 } & \ignore{ 86.51 } \\
    \bottomrule
    \end{tabular}
     }
    \fontsize{8}{9}\selectfont
    \caption{Comparison between the traditional stroke extraction method in~\cite{xu2016decomposition} and our stroke instance segmentation approach via accuracy, precision, recall and F1. $K_s$ and $H_s$ are the \textbf{s}ubsets with 100 randomly sampled datapoints from $K$ and $H$, respectively. $K_s^{*}$ denotes the 100 datapoints with the most strokes in $K$.
    }
    \label{tab:compare_traditional}
\end{table}

\begin{figure}[!htb]
\centering
\includegraphics[width=1\linewidth]{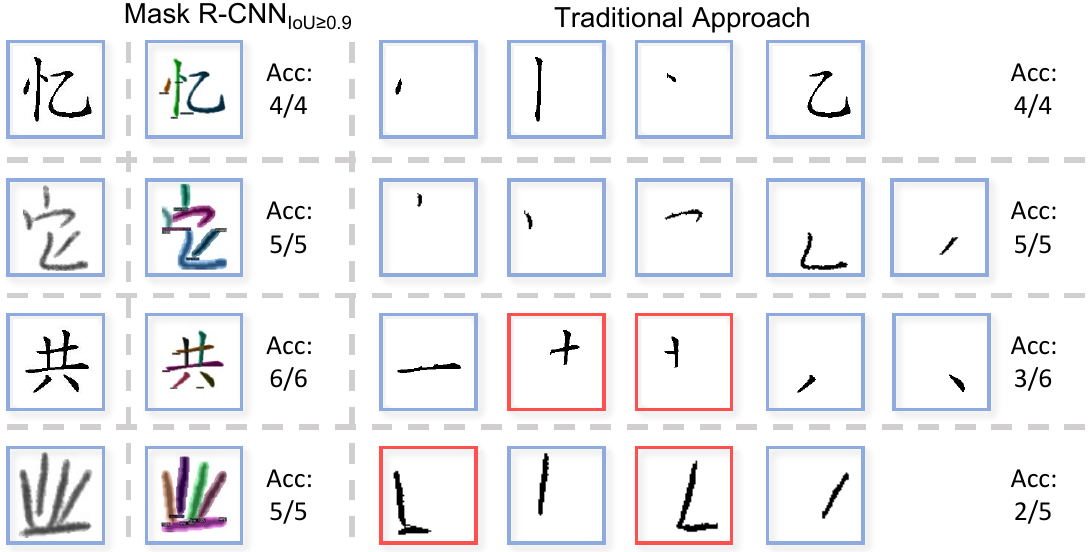}
\caption{Qualitative results from the method in~\cite{xu2016decomposition} and our stroke instance segmentation approach.}
\label{fig:cases_compare}
\end{figure}

\noindent{\textbf{Experiment Protocols}. Most of the previous approaches~\cite{sun2014geometric,xu2016decomposition} can \textit{only} deliver results on extracted stroke locations without corresponding categories. In this way, with no access to external databases, they can only benchmark their results on 100 images with human evaluation~\cite{sun2014geometric}. Specifically, given the extracted stroke images, a human is required to evaluate whether the extracted results contain the desired strokes. Then, accuracy is used as the evaluation metric\footnote{\lkyy{More details are put in the supplementary.}}. We follow these protocols for fair comparison. We also provide results in terms of precision, recall and F1-score for a more comprehensive evaluation.}

\subsubsection{Quantitative Results}
\alec{We report comparisons between the latest traditional stroke extraction method~\cite{xu2016decomposition} and our stroke instance segmentation approach in Table~\ref{tab:compare_traditional}. Since the traditional approach produces a perfect location match when correctly recognizing a stroke, we set a high IoU threshold \ie~0.9 and the extracted stroke that has a IoU overlap with GT higher than 0.9 is considered correctly extracted for fair comparisons. We have the following observations: \textbf{First}, the traditional approach performs worse on $K_s^{*}$ than $K_s$, indicating their limitations in handling characters with complex structures. \textbf{Second}, the traditional method is hard to recognize the character in handwritten dataset $H_s$ than \texttt{Kai Ti} dataset $K_s$, showing that the stroke with high variance poses a nontrivial challenge for this task. \textbf{Last}, on all datasets (\ie~$K_s^{*}$, $K_s$ and $H_s$), our stroke instance segmentation approach surpasses the previous method by a large margin under all metrics. Note that 0.9 is a very high IoU threshold in standard instance segmentation literature~\cite{he2017mask,wu2019detectron2}. As we lower the IoU threshold, we observe more significant gains. 
Improving the stroke instance segmentation performance under a high IoU threshold is a challenging mission to solve.}

\subsubsection{Qualitative Results}

\alec{We provide qualitative comparisons between the traditional approach~\cite{xu2016decomposition} and our method in Figure~\ref{fig:cases_compare}. We observe that: 1) Traditional approach can extract well-separable and regular strokes (row 1-2 in Figure~\ref{fig:cases_compare}). 2) It is very hard for them to extract strokes from characters with complex structures or with irregular shapes (rows 3-4 in Figure~\ref{fig:cases_compare}). Unlike them, we can well handle these cases, demonstrating the efficacy of the proposed datasets and stroke instance segmentation approach.}

\begin{table}[!t]
    \centering
    \resizebox{.7\linewidth}{!}{
    \begin{tabular}{c|ll}
    \toprule
     Pretrained & IoU ($\uparrow$) & MAE ($\downarrow$) \\
    \midrule
    \XSolidBrush & { 34.94 } & { 0.137 } \\
    ImageNet & { 43.04 } { \imporve{+8.10} } & { 0.117 } { \imporve{-0.020} } \\
    \ckcsdshort & { 44.21 } { \imporve{+9.27} } & { 0.114 } { \imporve{-0.023} } \\
    \chcsdshort & \textbf{ 44.84 } \textbf{ \imporve{+9.90} } & \textbf{ 0.112 } \textbf{ \imporve{-0.025} } \\
    \bottomrule
    \end{tabular}
    }
    \fontsize{8}{9}\selectfont
    \caption{Experiments on font generation task with different pretrained datasets.}
    \label{tab:font_generation}
\end{table}

\begin{table}[!h]
    \centering
    \resizebox{.85\linewidth}{!}{
    \begin{tabular}{l|c|ll}
    \toprule
     Eval. Type & Pretrained & Acc. ($\uparrow$) & MAE ($\downarrow$)  \\
    \midrule
    \multirow{3}{*}{End to End} & \XSolidBrush & { 63.03 } & { 16.92 } \\
                                & ImageNet & { 69.94 } { \imporve{+6.91} } & { 12.77 } { \imporve{-4.15} } \\
                                & \chcsdshort & \textbf{ 70.91 } \textbf{ \imporve{+7.88} } & \textbf{ 12.02 } \textbf{ \imporve{-4.90} } \\
    \midrule
    \multirow{3}{*}{Linear Probe} & \XSolidBrush & { 60.00 } & { 17.34 } \\
                                & ImageNet & { 65.86 } { \imporve{+5.86} } & { 16.37 } { \imporve{-0.97} } \\
                                & \chcsdshort & \textbf{ 67.47 } \textbf{ \imporve{+7.47} } & \textbf{ 14.31 } \textbf{ \imporve{-3.03} } \\
    
    \bottomrule
    \end{tabular}
    }
    \fontsize{8}{9}\selectfont
    \caption{Experiments on handwritten aesthetic assessment task using different pretrained datasets.
    }
    \label{tab:handwritten_assessment}
\end{table}

\subsection{\alec{Transferring Features to Downstream Tasks}\footnote{\alec{More details and results are put in the supplementary.}}}

\subsubsection{\alec{Font Generation}}
\alec{We investigate whether our trained features can be transferred to the font generation task~\cite{jiang2019scfont,liu2021fontrl}. We conduct experiments using fontRL~\cite{liu2021fontrl}, which uses a stroke Bounding Box Network (BBoxNet) to put each stroke of a character in the desired position before character rendering. Hence, we use different pretrained models to initialize the BBoxNet and the results are shown in Table~\ref{tab:font_generation}. IoU and MAE are used to evaluate the structural alignment and the appearance difference between the generated font and the GT font respectively. Using the model pretrained on our datasets, we achieve better performance than other pretrained models, especially on IoU, showing that our pretrained model better understands the character structure to facilitate this task.}

\subsubsection{\alec{Handwritten Aesthetic Assessment}} 
\alec{We study this task~\cite{sun2015aesthetic} with different pretrained models. Given a handwritten character image, this task requires the model to output a classification result (from good, medium and bad) and a regression result (range from 0 to 150) to indicate the aesthetic level of the handwritten. We initialize the ResNet-50 with different pretrained models. Moreover, we also employ the linear probing protocol that freezes the pretrained models and trains the classification and regression layer only to further inspect the features' effectiveness. In Table~\ref{tab:handwritten_assessment}, the model pretrained with our \chcsdshort dataset performs much better than the model pretrained with ImageNet that has more than 1M images, showing that a compact dataset with domain-specific character structure knowledge is more suitable than a large-scale general vision dataset for the handwritten aesthetic assessment task.}

\section{Conclusion}
\lky{In this work, we propose the first large-scale \bname (\bnameshort) benchmark to improve stroke extraction task \lec{and facilitate further research}. \lec{To this end, we effortlessly harvest a large number of Chinese character images and provide stroke-level annotation for them to create \ckcsdshort and \chcsdshort datasets. The proposed datasets satisfy both character-level and stroke-level diversities for achieving promising stroke extraction. We carry out a series of analyses on the properties of the proposed datasets and point out their intrinsic difficulties. Last, we conduct extensive experiments with stroke instance segmentation models to analyze the influential factors in delivering promising results and show that pretraining the model with the proposed datasets benefits the downstream tasks. Our future works will focus on improving the stroke segmentation performance \alec{under strict IoU condition}.} 
}


{
\bibliography{aaai23}

\begin{thebibliography}{32}
\providecommand{\natexlab}[1]{#1}

\bibitem[{Arcelli and Di~Baja(1985)}]{arcelli1985width}
Arcelli, C.; and Di~Baja, G.~S. 1985.
\newblock A width-independent fast thinning algorithm.
\newblock \emph{{TPAMI}}, 7: 463--474.

\bibitem[{Bolya et~al.(2019)Bolya, Zhou, Xiao, and Lee}]{Bolya_2019_ICCV}
Bolya, D.; Zhou, C.; Xiao, F.; and Lee, Y.~J. 2019.
\newblock {YOLACT:} Real-Time Instance Segmentation.
\newblock In \emph{{ICCV}}, 9156--9165.

\bibitem[{Cai and Vasconcelos(2018)}]{CascadeRCNN_2018_CVPR}
Cai, Z.; and Vasconcelos, N. 2018.
\newblock Cascade {R-CNN:} Delving Into High Quality Object Detection.
\newblock In \emph{{CVPR}}, 6154--6162.

\bibitem[{Cao and Tan(2000)}]{cao2000model}
Cao, R.; and Tan, C.~L. 2000.
\newblock A model of stroke extraction from chinese character images.
\newblock In \emph{{ICPR}}, 368--371.

\bibitem[{Chen et~al.(2016)Chen, Lian, Tang, and Xiao}]{chen2016benchmark}
Chen, X.; Lian, Z.; Tang, Y.; and Xiao, J. 2016.
\newblock A benchmark for stroke extraction of chinese characters.
\newblock \emph{Acta Scientiarum Naturalium Universitatis Pekinensis}, 52:
  49--57.

\bibitem[{Chen et~al.(2017)Chen, Lian, Tang, and Xiao}]{chen2017automatic}
Chen, X.; Lian, Z.; Tang, Y.; and Xiao, J. 2017.
\newblock An Automatic Stroke Extraction Method using Manifold Learning.
\newblock In \emph{{Eurographics}}, 65--68.

\bibitem[{Fan and Wu(2000)}]{fan2000run}
Fan, K.-C.; and Wu, W.-H. 2000.
\newblock A run-length-coding-based approach to stroke extraction of Chinese
  characters.
\newblock \emph{{PR}}, 33: 1881--1895.

\bibitem[{Ganin and Lempitsky(2015)}]{ganin2015unsupervised}
Ganin, Y.; and Lempitsky, V. 2015.
\newblock Unsupervised domain adaptation by backpropagation.
\newblock In \emph{{ICML}}, 1180--1189.

\bibitem[{Gao and Wu(2020)}]{gao2020gan}
Gao, Y.; and Wu, J. 2020.
\newblock GAN-Based Unpaired Chinese Character Image Translation via Skeleton
  Transformation and Stroke Rendering.
\newblock In \emph{{AAAI}}, 646--653.

\bibitem[{He et~al.(2017)He, Gkioxari, Doll{\'a}r, and Girshick}]{he2017mask}
He, K.; Gkioxari, G.; Doll{\'a}r, P.; and Girshick, R. 2017.
\newblock Mask r-cnn.
\newblock In \emph{{ICCV}}, 2961--2969.

\bibitem[{Hsieh et~al.(2021)Hsieh, Robb, Chen, and Huang}]{hsieh2021droploss}
Hsieh, T.-I.; Robb, E.; Chen, H.-T.; and Huang, J.-B. 2021.
\newblock Droploss for long-tail instance segmentation.
\newblock In \emph{{AAAI}}, 1549--1557.

\bibitem[{Huang et~al.(2020)Huang, He, Jin, and Wang}]{huang2020rd}
Huang, Y.; He, M.; Jin, L.; and Wang, Y. 2020.
\newblock RD-GAN: few/zero-shot chinese character style transfer via radical
  decomposition and rendering.
\newblock In \emph{{ECCV}}, 156--172.

\bibitem[{Jiang et~al.(2019)Jiang, Lian, Tang, and Xiao}]{jiang2019scfont}
Jiang, Y.; Lian, Z.; Tang, Y.; and Xiao, J. 2019.
\newblock Scfont: Structure-guided chinese font generation via deep stacked
  networks.
\newblock In \emph{{AAAI}}, 4015--4022.

\bibitem[{Lee and Wu(1998)}]{lee1998chinese}
Lee, C.; and Wu, B. 1998.
\newblock A Chinese-character-stroke-extraction algorithm based on contour
  information.
\newblock \emph{{PR}}, 31: 651--663.

\bibitem[{Liu, Kim, and Kim(2001)}]{liu2001model}
Liu, C.-L.; Kim, I.-J.; and Kim, J.~H. 2001.
\newblock Model-based stroke extraction and matching for handwritten Chinese
  character recognition.
\newblock \emph{{PR}}, 34: 2339--2352.

\bibitem[{Liu, Jia, and Tan(2006)}]{liu2006geometrical}
Liu, X.; Jia, Y.; and Tan, M. 2006.
\newblock Geometrical-statistical modeling of character structures for natural
  stroke extraction and matching.
\newblock In \emph{{IWFHR}}.

\bibitem[{Liu and Lian(2021)}]{liu2021fontrl}
Liu, Y.; and Lian, Z. 2021.
\newblock FontRL: Chinese Font Synthesis via Deep Reinforcement Learning.
\newblock In \emph{{AAAI}}, 2198--2206.

\bibitem[{Qiguang(2004)}]{qiguang2004algorithm}
Qiguang, L. Z.~H. 2004.
\newblock Algorithm and implementation in chinese charac-tersorder of strokes
  recognition.
\newblock \emph{{CAS}}, 7: 041.

\bibitem[{Ren et~al.(2015)Ren, He, Girshick, and Sun}]{ren2015faster}
Ren, S.; He, K.; Girshick, R.~B.; and Sun, J. 2015.
\newblock Faster {R-CNN:} Towards Real-Time Object Detection with Region
  Proposal Networks.
\newblock 91--99.

\bibitem[{Su and Wang(2004)}]{su2004decomposing}
Su, Y.-M.; and Wang, J.-F. 2004.
\newblock Decomposing Chinese characters into stroke segments using SOGD
  filters and orientation normalization.
\newblock In \emph{{ICPR}}, 351--354.

\bibitem[{Su, Cao, and Wang(2009)}]{su2009stroke}
Su, Z.; Cao, Z.; and Wang, Y. 2009.
\newblock Stroke extraction based on ambiguous zone detection: a preprocessing
  step to recover dynamic information from handwritten Chinese characters.
\newblock \emph{{IJDAR}}, 12: 109--121.

\bibitem[{Sun et~al.(2015)Sun, Lian, Tang, and Xiao}]{sun2015aesthetic}
Sun, R.; Lian, Z.; Tang, Y.; and Xiao, J. 2015.
\newblock Aesthetic Visual Quality Evaluation of Chinese Handwritings.
\newblock In \emph{{IJCAI}}, 2510--2516.

\bibitem[{Sun, Qian, and Xu(2014)}]{sun2014geometric}
Sun, Y.; Qian, H.; and Xu, Y. 2014.
\newblock A geometric approach to stroke extraction for the Chinese calligraphy
  robot.
\newblock In \emph{{ICRA}}, 3207--3212.

\bibitem[{Tian et~al.(2019)Tian, Shen, Chen, and He}]{tian2019fcos}
Tian, Z.; Shen, C.; Chen, H.; and He, T. 2019.
\newblock Fcos: Fully convolutional one-stage object detection.
\newblock In \emph{{ICCV}}, 9627--9636.

\bibitem[{Wu et~al.(2019)Wu, Kirillov, Massa, Lo, and
  Girshick}]{wu2019detectron2}
Wu, Y.; Kirillov, A.; Massa, F.; Lo, W.-Y.; and Girshick, R. 2019.
\newblock Detectron2.
\newblock \url{https://github.com/facebookresearch/detectron2}.

\bibitem[{Xie et~al.(2021)Xie, Chen, Sun, and Lu}]{xie2021dg}
Xie, Y.; Chen, X.; Sun, L.; and Lu, Y. 2021.
\newblock DG-Font: Deformable Generative Networks for Unsupervised Font
  Generation.
\newblock In \emph{{CVPR}}, 5130--5140.

\bibitem[{Xu et~al.(2007)Xu, Jiang, Lau, and Pan}]{xu2007intelligent}
Xu, S.; Jiang, H.; Lau, F. C.-M.; and Pan, Y. 2007.
\newblock An intelligent system for chinese calligraphy.
\newblock In \emph{{AAAI}}, 1578--1583.

\bibitem[{Xu et~al.(2016)Xu, Liang, Zhang, Dong, and
  Izquierdo}]{xu2016decomposition}
Xu, Z.; Liang, Y.; Zhang, Q.; Dong, L.; and Izquierdo, E. 2016.
\newblock Decomposition and matching: Towards efficient automatic Chinese
  character stroke extraction.
\newblock In \emph{{VCIP}}, 1--4.

\bibitem[{Xun et~al.(2015)Xun, Xiaochen, Weihua, Sun, and Ramp}]{xun2015stroke}
Xun, E.; Xiaochen, L.; Weihua, A.; Sun, Y.; and Ramp, I. 2015.
\newblock Stroke retrieval of handwritten Chinese character images for
  handwriting teaching.
\newblock \emph{Scientiarum Naturalium Universitatis Pekinensis}, 51: 241--248.

\bibitem[{Yu, Wu, and Yuan(2012)}]{yu2012stroke}
Yu, K.; Wu, J.; and Yuan, Z. 2012.
\newblock Stroke extraction for chinese calligraphy characters.
\newblock \emph{{JCIS}}, 8: 2493--2500.

\bibitem[{Zeng et~al.(2021)Zeng, Chen, Liu, Wang, and Yao}]{zeng2021strokegan}
Zeng, J.; Chen, Q.; Liu, Y.; Wang, M.; and Yao, Y. 2021.
\newblock Strokegan: Reducing mode collapse in Chinese font generation via
  stroke encoding.
\newblock In \emph{{AAAI}}, 3270--3277.

\bibitem[{Zeng et~al.(2010)Zeng, Feng, Xie, and Liu}]{zeng2010cascade}
Zeng, J.; Feng, W.; Xie, L.; and Liu, Z.-Q. 2010.
\newblock Cascade Markov random fields for stroke extraction of Chinese
  characters.
\newblock \emph{{IS}}, 180: 301--311.

\end{thebibliography}
}

\clearpage


\section*{Supplementary material}

# Instance Segmentation for Chinese Character Stroke Extraction, Datasets and Benchmarks (Supplementary Materials)

**Lizhao Liu**[1*], **Kunyang Lin**[1], **Shangxin Huang**[1], **Zhongli Li**[2],
**Chao Li**[3], **Yunbo Cao**[2], **and Qingyu Zhou**[2†]

[1] South China University of Technology, [2] Tencent Cloud Xiaowei, [3] Xiaomi Group,
selizhaoliu@mail.scut.edu.cn, qingyuzhou@tencent.com

We organize our supplementary materials as follows.

- In Section "More Analysis on Stroke Detection", we provide a detailed analysis of the stroke detection process.

- In Section "More Transferability Results on Standard Fonts", we provide more qualitative results on other standard fonts.

- In Section "More Failure Cases", we provide more failure cases on the stroke instance segmentation.

- In Section "More Details on the Previous Methods", we provide more details on previous stroke extraction approach(Xu et al. 2016).

- In Section "More Details on the Downstream Tasks", we provide more experimental details on the downstream tasks, *i.e.,* experimental settings, evaluation metrics and more results.

- In Section "More Results on the Background Effect", we provide more qualitative results on the effect of the background.

## More Analysis on Stroke Detection

Since we adopt the Mask R-CNN (He et al. 2017) model for stroke instance segmentation, which first performs stroke detection and then segmentation, we analyze the stroke detection process of Faster R-CNN (Ren et al. 2015) (the detection part of Mask R-CNN) process by tuning important hyper-parameters: the number of stages, anchor box, backbone, and image resolution.

**Effect of the Number of Stages** In this section, we provide experiments on widely used one-stage, two-stage as well as multiple-stage detectors to see how they perform on our proposed datasets. The results are in Table 1. Note that we set the number of stages of Cascade R-CNN to 3, resulting in a 4-stage model (RPN contributes 1 stage). The results conclude that as the number of stages increases, we obtain better stroke detection results.

---

[*]This work was partially done while the author was an intern at Tencent Cloud Xiaowei.
[†]Corresponding author.

| Detector | #Stage | $AP^{box}_{50}$ | $AP^{box}_{75}$ | $AP^{box}$ |
|---|---|---|---|---|
| FCOS | One | 88.24 | 79.18 | 69.52 |
| Faster R-CNN | Two | **90.83** | 82.48 | 71.07 |
| Cascade R-CNN | Multiple | 88.39 | **83.58** | **74.15** |

Table 1: Stroke detection results of one-stage, two-stage, and multiple-stage detectors.

**Effect of the Anchor Box** As shown by previous researches (Ren et al. 2015), the ratios of the anchor box have a great effect on the final results. Due to the strip shape property of the stroke, we analyse how the choices of anchor box affect the stroke detection performance. Specifically, we gradually add anchor boxes with higher ratios into Faster R-CNN. The results are in Table 2, which conveys that incorporating anchor boxes with higher ratio indeed improve the stroke detection performance.

| Anchor Ratios | $AP^{box}_{50}$ | $AP^{box}_{75}$ | $AP^{box}$ |
|---|---|---|---|
| $S_1 = \{1{:}1\}$ | 89.91 | 81.37 | 69.12 |
| $S_2 = S_1 \cup \{1{:}2, 2{:}1\}$ | 90.19 | **80.87** | 69.60 |
| $S_3 = S_2 \cup \{1{:}4, 4{:}1\}$ | 88.24 | 79.18 | 69.52 |
| $S_4 = S_3 \cup \{1{:}8, 8{:}1\}$ | **90.75** | 80.33 | **71.03** |

Table 2: Stroke detection results in different anchor ratios.

**Effect of the Image Resolution** Currently, most of the detectors are evaluated on COCO (Lin et al. 2014) benchmark that has a relatively large image resolution *i.e.,* (800∼1000). However, the resolution of Chinese character images is typically small *i.e.,* (80∼120). It is not clear whether the detectors are biased by the image resolution. Thus, we provide experiments on different resolutions to see their effect. Specifically, we apply the equal scaling strategy while resizing the short size into one of $\{112, 120\}$, $\{224, 240\}$ and $\{448, 480\}$ to see the performance of the Faster R-CNN detector. Note that we also accordingly scale the anchor boxes. In Table 3, we achieve the best $AP^{\textbf{box}}$ at resolution $\{448, 480\}$ but also bring higher computation cost.

**Effect of the Backbone** We provide experiments on different backbone settings for Faster R-CNN to show how they affect the final performance. Specifically, we consider two settings *e.g.,* different backbone architectures and pretrained

or not on ImageNet (Deng et al. 2009). For backbone architecture, we use ResNet-{50,101} (He et al. 2016) and FPN-{50,101} (Lin et al. 2017). Table 4 shows the experiment results, which show that pretraining and deeper models boost the performance.

## More Transferability Results on Standard Fonts

In Figure 1, we show more transferability results on the other standard font styles. Specifically, we leverage the model trained by our CCSE-Kai dataset to automatically label character images of other font styles. Most of the results share the competitive performance as the result of the source font style (*i.e.,* Kai Ti). We attribute this strong performance to the highly similar stroke styles shared by these font styles. We notice that in some cases the stroke extraction other than Kai Ti are inaccurate. We believe this could be addressed by weakly supervised instance segmentation approach (Zhou et al. 2018) since the stroke categories and the stroke composition of each character are identical across different font styles. We leave it to future work.

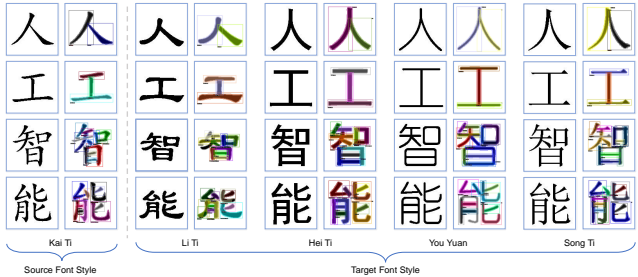

Figure 1: Stroke extraction results for Kai Ti, Li Ti, Hei Ti, You Yuan, Song Ti styles with the model trained on the Kai Ti dataset. Best view with zoom-in.

## More Failure Cases

We analyze more failure cases in our stroke instance segmentation model. In Figure 2, we observe that the following situations may lead to the failure of stroke extraction: (1) Two separate strokes are connected and thus similar to another stroke. As shown in Figure 2a, the connection between shu and heng happens to be similar to heng zhe; (2) Missed detection caused by the bad hyper-parameter choices such as high confidence score (Figure 2b); (3) Some strokes are so similar that they are sometimes indistinguishable, such as shu wan and shu ti in Figure 2c; (4) One stroke is detected as multiple strokes, as shown in Figure 2d;

| Resolutions | $AP_{50}^{box}$ | $AP_{75}^{box}$ | $AP^{box}$ |
|---|---|---|---|
| {112, 120} | 90.83 | **82.48** | 71.07 |
| {224, 240} | 90.20 | 80.72 | 72.14 |
| {448, 480} | **90.68** | 81.36 | **72.74** |

Table 3: Experiments on different image resolutions.

| Backbone | Depth | Pretrained | $AP_{50}^{box}$ | $AP_{75}^{box}$ | $AP^{box}$ |
|---|---|---|---|---|---|
| ResNet | 50 | × | 86.10 | 67.93 | 58.03 |
| ResNet | 50 | ✓ | **90.63** | **81.30** | **71.15** |
| FPN | 50 | × | 84.59 | 63.72 | 55.14 |
| FPN | 50 | ✓ | **90.83** | **82.48** | **71.07** |
| ResNet | 101 | ✓ | 88.70 | 81.86 | 71.05 |
| FPN | 101 | ✓ | 90.00 | 81.04 | 70.90 |

Table 4: Stroke detection results with different backbones and whether pretrained.

(5) The stroke has a thin and long shape, as shown in Figure 2e; (6) A small fraction of one stroke is falsely detected due to the occlusion by another stroke, as seen in Figure 2f. We believe these issues can be resolved by incorporating the stroke instance segmentation model with more character structure information and the occurrence relationship between strokes.

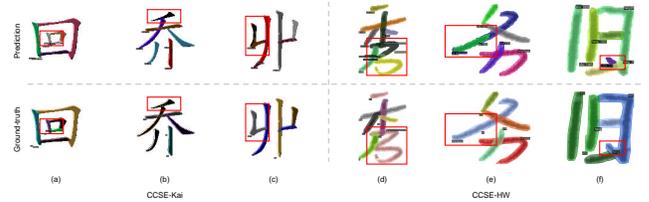

Figure 2: More failure cases. The first row is the stoke instance segmentation results and the second row is the ground truth. Best view with zoom-in.

## More Details on the Previous Methods

We provide the workflow of the latest traditional stroke extraction method ACSE (Xu et al. 2016) in Figure 3, which consists of three steps: character decomposition, cross area extraction and slope-based stroke combination. In ACSE, the character is first decomposed into several independent components according to the connectivity. Then, the skeletons and contours are extracted to compute the cross point sets and end point sets. If a skeleton's cross point set is empty, the corresponding stroke will be directly output as a simple stroke. According to the cross point set and end point set, each component is departed into several stroke segments. Last, several stroke segments are combined into one stroke if they share a similar slope.

## More Details on the Downstream Tasks

In this section, we provide more details on using the models pretrained on the proposed CCSE-Kai and CCSE-HW datasets to the downstream font generation and handwritten aesthetic assessment tasks.

### Font Generation

**Dataset**. Following the previous approach (Liu and Lian 2021), we conduct the font generation task on the digital handwritten font FZJHSXJW. The training set includes 775 Chinese characters with 7,004 strokes. The test set is consist

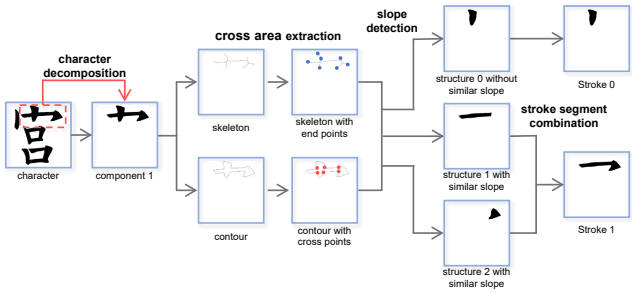

Figure 3: The overall workflow of the previous method (Xu et al. 2016). Using only the decomposed component 1 for simplicity.

of 6,763 Chinese characters with annotated stroke skeletons and the corresponding mean font styles. For evaluation, both the generated font and GT font are first converted into a binary mask with 1 denoting the pixel with stroke and 0 denoting the background. Then the Intersection over Union (IoU) and Mean Absolute Error (MAE) between the generated and GT fonts are used as the metrics

**Implementation details**. We leverage the fontRL in (Liu and Lian 2021) as our baseline method to test font generation performance with different pretrained datasets. The font generation process is as follows: given a stroke trajectory in a mean font style, 1) a Modification Parameter Network (MPNet) is applied to bend it into the stroke trajectory in a target font style; 2) a Bounding Box predicting Network (BBoxNet) is used to predict the location of the bent stroke in a canvas; 3) the above process is repeated for all strokes to form the complete skeleton of the target font; 4) an Image Rendering Module (IRM) is final employed to convert the complete skeleton into a glyph font image in an image-to-image translation manner. The MPNet, BBoxNet and IRM are trained sequentially. We conduct experiments by initializing the BBoxNet with parameters pretrained from different datasets: None, ImageNet (Deng et al. 2009), our CCSE-Kai and CCSE-HW while keeping other modules identical. The L1 loss is used as loss function to train BBoxNet. For optimization, we use Adam optimizer with lr $= 1e^{-4}$ for the first 100 epochs and lr $= 1e^{-5}$ for the last 50 epochs. Other experiment protocols are strictly aligned with (Liu and Lian 2021).

**Results**. We provide the evolution of the training process w.r.t. training loss, test IoU and test MAE in Figure 4. We have the following observations: **First**, with no dataset to pretrain, the convergence of the training loss is slow and very unstable. **Second**, all models with pretrained dataset is able to provide fast but not necessarily stable (*i.e.,* ImageNet) convergence. **Third**, models pretrained with our CCSE-Kai and CCSE-HW not only provide fast and stable convergence but also boost the IoU metric considerably. This indicates that our datasets with 10K images can beat the much larger ImageNet dataset with 1M images in this task by providing effective character structure information instead of the general visual features.

**Handwritten Aesthetic Assessment**

**Dataset**. We use the Chinese Handwriting Aesthetic Evaluation Dataset (CHAED) from (Sun et al. 2015), which consists of 1000 images, 10 for every 100 Chinese characters. Odd-numbered and even-numbered images are used for training and testing, respectively. Each image is labeled by 33 people, selecting one from 3 levels: good, medium and bad. For each image, denote the number of people who label it to good, medium and bad by $p_{\text{good}}, p_{\text{medium}}$ and $p_{\text{bad}}$. The final classification label is computed by $\arg\max_i p_i, i \in \{\text{good}, \text{medium}, \text{bad}\}$. Moreover, the aesthetic score is calculated by $100 \times \text{good} + 50 \times \text{medium} + 0 \times \text{bad}$.

**Implementation details**. As for the model architecture, we use ResNet-50 as feature extractor with different pretrained datasets *i.e.,* None, ImageNet (Deng et al. 2009), and our CCSE-HW. We drop the original 1000-class classification layer of the model pretrained in ImageNet or None. We drop the FCN detector head and segmentation head in Mask R-CNN after pretrained in CCSE-HW. In this way, the ResNet-50 serves to produce feature vectors in $\mathbb{R}^{2048}$. The last classification layer and regression layer are randomly initialized fully-connected layers. We leverage the earth mover distance loss (Talebi and Milanfar 2018) and smooth-l1 loss as the objective function for the aesthetic classification task and the regression task, respectively. For optimization, we use SGD optimizer with (lr, weight decay, momentum) $= (5e^{-3}, 1e^{-2}, 9e^{-1})$. We train all models for 120 epochs and the learning rate is decay by 0.1 every 40 epochs. We use batch size = 64. For image augmentation, random five crops and centre crop to size $224 \times 224$ are applied during training and testing, respectively.

**Results**. We provide the evolution of end-to-end training process w.r.t. classification training loss, regression training loss, test accuracy and test mean absolute error (MAE) in Figure 5. Compared to models pretrained on None and ImageNet (Deng et al. 2009), the model pretrained on our CCSE-HW converges more stable and much faster, achieving strong results over them in test accuracy and MAE. These results verify that pretraining the model using our dataset is able to benefit the downstream handwritten aesthetic assessment task, demonstrating the effectiveness of our datasets and the proposed stroke instance segmentation model. We believe there is a more effective way to improve the downstream tasks and we leave it to future works.

## More Results on the Background Effect

In order to evaluate the effect of the background added to CCSE-Kai and CCSE-HW, we conduct more experiments. We leverage the models trained with the pure datasets and the complex background augmented datasets to evaluate their performance on the noisy images. In Figure 6, when using the models trained on the pure CCSE-Kai and CCSE-HW, we unsurprisingly observe that there are a lot of false detections on the complex background. We speculate this is because the complex background contains stroke-like structures and some color interference, which was not taken into account during the training stage. To compensate for this, we propose to train the model with complex background aug-

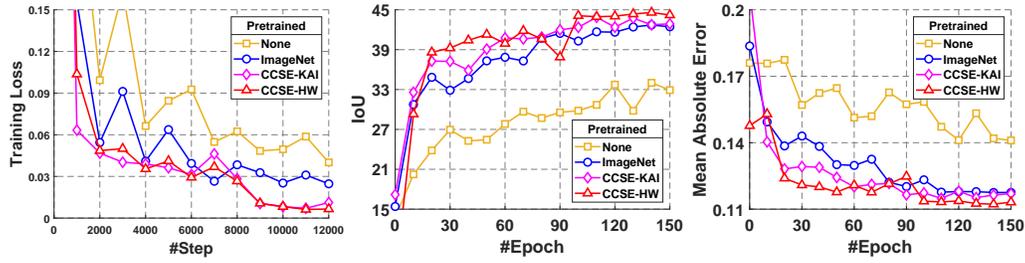

Figure 4: Results on font generation task using different pretrained models. From left to right: (a) training loss, (b) test intersection over union and (c) test mean absolute error.

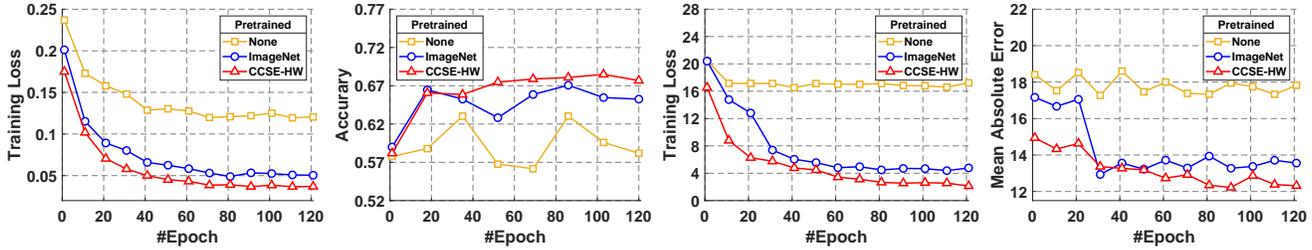

Figure 5: Results on handwritten aesthetic assessment task using different pretrained models. From left to right: (a) training loss on aesthetic classification task, (b) test accuracy on aesthetic classification task, (c) training loss on aesthetic regression task and (d) test mean absolute error on aesthetic regression task.

mented images. As shown in Figure 6, the performance is boosted substantially. It can be seen that the model can significantly reduce false detections. This further demonstrates that our datasets are applicable to the real world with minor efforts.

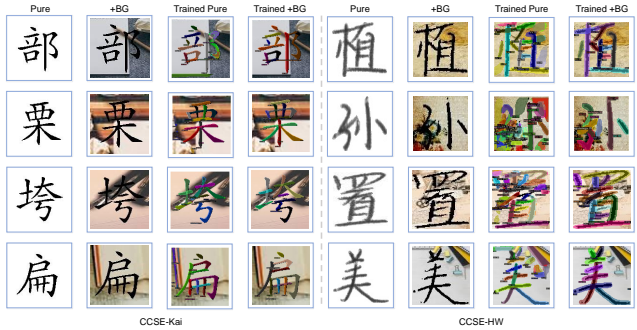

Figure 6: From left to right, "Pure": no complex background added. "+BG": add complex background. "Trained-Pure": the model used for inference is trained with pure datasets. "Trained +BG": the model used for inference is trained with the complex background augmented datasets. Best view with zoom-in.

## References

Deng, J.; Dong, W.; Socher, R.; Li, L.-J.; Li, K.; and Fei-Fei, L. 2009. Imagenet: A large-scale hierarchical image database. In *CVPR*, 248–255.

He, K.; Gkioxari, G.; Dollár, P.; and Girshick, R. 2017. Mask r-cnn. In *ICCV*, 2961–2969.

He, K.; Zhang, X.; Ren, S.; and Sun, J. 2016. Deep residual learning for image recognition. In *CVPR*, 770–778.

Lin, T.-Y.; Dollár, P.; Girshick, R.; He, K.; Hariharan, B.; and Belongie, S. 2017. Feature pyramid networks for object detection. In *CVPR*, 2117–2125.

Lin, T.-Y.; Maire, M.; Belongie, S.; Hays, J.; Perona, P.; Ramanan, D.; Dollár, P.; and Zitnick, C. L. 2014. Microsoft coco: Common objects in context. In *ECCV*, 740–755.

Liu, Y.; and Lian, Z. 2021. FontRL: Chinese Font Synthesis via Deep Reinforcement Learning. In *AAAI*, 2198–2206.

Ren, S.; He, K.; Girshick, R. B.; and Sun, J. 2015. Faster R-CNN: Towards Real-Time Object Detection with Region Proposal Networks. 91–99.

Sun, R.; Lian, Z.; Tang, Y.; and Xiao, J. 2015. Aesthetic Visual Quality Evaluation of Chinese Handwritings. In *IJCAI*, 2510–2516.

Talebi, H.; and Milanfar, P. 2018. NIMA: Neural image assessment. *TIP*, 27: 3998–4011.

Xu, Z.; Liang, Y.; Zhang, Q.; Dong, L.; and Izquierdo, E. 2016. Decomposition and matching: Towards efficient automatic Chinese character stroke extraction. In *VCIP*, 1–4.

Zhou, Y.; Zhu, Y.; Ye, Q.; Qiu, Q.; and Jiao, J. 2018. Weakly supervised instance segmentation using class peak response. In *CVPR*, 3791–3800.



\end{document}